\documentclass[Afour,sagev,times]{sagej}

\usepackage{moreverb,url}

\usepackage{booktabs}
\usepackage{multirow}

\usepackage{float}

\usepackage[colorlinks,
            bookmarksopen,
            bookmarksnumbered,
            citecolor=red,
            urlcolor=red, 
            unicode]{hyperref}

\usepackage[absolute,overlay]{textpos}
\usepackage{pbox}
\usepackage{everypage}

\newcommand{\stamp}[1][© 2023 SAGE. This is the author's version of the article that has been published in Information Visualization. The final version of this record is available at: \href{https://doi.org/10.1177/14738716221142005}{\color{blue}10.1177/14738716221142005}]{%
\begin{textblock*}{140mm}(37mm,279mm)
\centering%
\small%
\emph{#1}%
\end{textblock*}%
}

\AddEverypageHook{
  \stamp
}

\graphicspath{{figures/}{pictures/}{images/}{./}} 

\usepackage{booktabs,ragged2e}

\usepackage[flushleft]{threeparttable}

\usepackage{amssymb}

\usepackage{caption}
\usepackage{booktabs}
\usepackage{multirow}

\usepackage{amsmath}
\usepackage{algorithm}
\newcommand{\euler}{e}

\usepackage{mathptmx}                  

\usepackage[dvipsnames]{xcolor} 
\newcommand{\hl}[1]{#1} 

\newcommand{\circled}[1]{\raisebox{.4pt}{\textcircled{\raisebox{-.0pt} {\tiny \bf  #1}}}}

\newcommand\BibTeX{{\rmfamily B\kern-.05em \textsc{i\kern-.025em b}\kern-.08em
T\kern-.1667em\lower.7ex\hbox{E}\kern-.125emX}}

\def\volumeyear{2022}

\usepackage{times}
\usepackage{graphicx}

\usepackage[super]{nth}
\usepackage{csquotes}

\newlength{\boxh}
\begin{document}

\runninghead{Chatzimparmpas \textit{et~al.}}

\title{VisRuler: Visual Analytics for Extracting Decision Rules from Bagged and Boosted Decision Trees}

\author{Angelos Chatzimparmpas\affilnum{1}, Rafael M.\ Martins\affilnum{1}, and Andreas Kerren\affilnum{1, 2}}

\affiliation{\affilnum{1}Linnaeus University, Sweden\\
\affilnum{2}Linköping University, Sweden}

\corrauth{Angelos Chatzimparmpas,
Department of Computer Science and Media Technology,
Linnaeus University,
351 95 Växjö,
Sweden.}

\email{angelos.chatzimparmpas@lnu.se}


\begin{abstract}
Bagging and boosting are two popular ensemble methods in machine learning (ML) that produce many individual decision trees. Due to the inherent ensemble characteristic of these methods, they typically outperform single decision trees or other ML models in predictive performance. However, numerous decision paths are generated for each decision tree, increasing the overall complexity of the model and hindering its use in domains that require trustworthy and explainable decisions, such as finance, social care, and health care. Thus, the interpretability of bagging and boosting algorithms---such as random forest and adaptive boosting---reduces as the number of decisions rises. In this paper, we propose a visual analytics tool that aims to assist users in extracting decisions from such ML models via a thorough visual inspection workflow that includes selecting a set of robust and diverse models (originating from different ensemble learning algorithms), choosing important features according to their global contribution, and deciding which decisions are essential for global explanation (or locally, for specific cases). The outcome is a final decision based on the class agreement of several models and the explored manual decisions exported by users. We evaluated the applicability and effectiveness of VisRuler via a use case, a usage scenario, and a user study. 
The evaluation revealed that most users managed to successfully use our system to explore decision rules visually, performing the proposed tasks and answering the given questions in a satisfying way.
\end{abstract}


\stamp

\maketitle

\section{Introduction} \label{sec:intro}
  Ensemble learning (EL)~\cite{Zhou2009Ensemble} is a well-established area of machine learning (ML) that strives for better performance by merging the predictions from various ML models.
Three prominent methods for building ensembles are:~\cite{Sagi2018Ensemble}
%
bagging,~\cite{Breiman1996Stacked} boosting,~\cite{Freund1996Experiments,Schapire1990Strength} and stacking.~\cite{Wolpert1992Stacked} 
Bagging requires training many decision trees on separate groups of instances and taking the average of their predictions.~\cite{Breiman1996Stacked}
Boosting attaches weak classifiers (e.g., decision stumps or shallow decision trees) sequentially, each improving the predictions made by the previous models.~\cite{Freund1996Experiments,Schapire1990Strength} 
Stacking involves fitting many base models from different algorithms on the same data set and using a metamodel to combine their results.~\cite{Wolpert1992Stacked} The common ground between bagging and boosting methods is that they incorporate ML algorithms that produce numerous decision trees,~\cite{Kingsford2008Decision} such as random forest (RF)~\cite{Breiman2001Random} and adaptive boosting/AdaBoost (AB),~\cite{Freund1999A} respectively. The decision paths stemming from bagged or boosted decision trees are the target of the visual analytics (VA) approach proposed in this paper.

The popularity of RF and AB is confirmed by their success in solving typical supervised classification problems, which constitute the majority of problems in the real world.~\cite{Opitz1999Popular,Wyner2017Explaining} An in-depth study~\cite{Delgado2014Do} that estimates the performances of 179 algorithms of various types~\cite{Dua2017} concludes that bagged decision trees of RF are better than other (types of) algorithms, such as deep learning approaches. 
%
Despite their remarkable predictive power, a crucial concern for algorithms that generate many decision trees is \emph{interpretability}. Brieman,~\cite{Breiman2001Statistical} for instance, indicates that RF models, while superb predictors, receive a low rating regarding their interpretability. As ML models can provide incorrect predictions,~\cite{Caruana2015Intelligible} ML experts have to check whether the model functions properly.~\cite{Tam2017An} Also, domain experts in critical fields need to understand how a specific prediction has been reached in order to trust in ML.~\cite{Zhou20182D} For example, in medicine, a physician might not rely on a model without explanations of how and why it forms a prediction, since patient lives are at risk.~\cite{Ribeiro2016Why,Hastie2001The,Lakkaraju2016Interpretable} Or, in the financial domain, declined decisions for loan applicants require additional transparency with the precise justification of the outcome.~\cite{Sachan2020An}
Although both algorithms follow the same concept of growing decision trees, their objectives differ: AB focuses on correcting misclassified training instances, while RF mainly reduces variance to achieve better generalizability. However, this fundamental goal of the former makes it susceptible to noisy cases,~\cite{Bauer1999Empirical} while the latter arguably remains intact.~\cite{Kotsiantis2007Combining}
Thus, one research question that remains open is: 
%
%
%
\hl{\textbf{(RQ1)} Given the differences between rules obtained from bagged decision trees and those derived from boosted decision trees, how does their combination lead to potential benefits, regarding interpretability enhancement for decision making?}
%

The interpretation of ML models typically happens either at a global or a local level.~\cite{Kopitar2019Local}
Global approaches intend to explain the ML model as a whole,~\cite{Lipton2018The} assisting domain experts in exploring the general impact of each decision and gaining confidence in the produced predictions. 
On the other hand, local approaches aim to provide case-based reasoning,~\cite{Du2019Techniques,Carvalho2019Machine} allowing domain experts to review a prediction and trace its decision path in order to conclude if the decision rule, and consequently the prediction, is trustworthy.~\cite{Weller2019Transparency} 
Nevertheless, comparing numerous alternative decision paths without the support of an intelligent system is a time-consuming and resource-heavy procedure. For example, to scan the list of test instances rapidly and investigate specific instances of interest from multiple perspectives (e.g., outliers and borderline cases) can be crucial.~\cite{Kim2014The} Thus, the whole process can benefit from a fast approach for automatic generation and semi-automatic exploration of reliable decisions with ML experts' involvement.
It should also result in robust decisions, since domain experts are the most suitable for carefully examining and then manually picking sensible decisions according to their prior experience and understanding. One research question that arises from this (possibly under-researched) need for cooperation, starting from the selection of models to the extraction of insightful decisions, is: \textbf{(RQ2)} How can VA tools/systems support the collaboration between ML experts and domain experts?

In this paper, we present \textsc{VisRuler},
a VA tool that addresses the research questions described above by supporting the exploratory combination of decisions from two closely-related ML algorithms (i.e., RF and AB). \textsc{VisRuler} uses validation metrics for picking performant and diverse models and combines the decision paths from bagged and boosted trees to extract insightful and interpretable rules.
Our contributions consist of the following:

\begin{itemize}
\item a visual analytic workflow for defining a methodical way of evaluating decisions (cf. Figure~\ref{fig:workflow-diagram} described in Section~\nameref{sec:overview});
\item a prototype VA tool, called \textsc{VisRuler}, that applies the suggested workflow with coordinated views that support the joint effort between ML experts and domain experts for extracting rules and making decisions, respectively;
\item a use case and a usage scenario, applying real-world data, that validate the effectiveness of utilizing both bagged and boosted decision trees at the same time; and
\item a user study that showed promising results.
\end{itemize}      

\noindent The rest of this paper is organized as follows. In Section~\nameref{sec:relwo}, we discuss relevant techniques for visualizing bagging and boosting decision trees, along with tree- and rule-based models and a bulk of relevant works of visual analytics systems for multi-model comparison. Section~\nameref{sec:back} explains the core differences between bagging and boosting, and it further motivates why mixing decisions stemming from both algorithms could be beneficial for the users. In Section~\nameref{sec:goals}, we describe the design goals and analytical tasks for comparing alternative decision rules, and we present the target groups (i.e., our stakeholders). Section~\nameref{sec:overview} focuses on the functionalities of the tool and describes the first use case with the goal of identifying which countries have a higher happiness score index and why. Next, in Section~\nameref{sec:case}, we demonstrate the applicability and usefulness of \textsc{VisRuler} with a usage scenario comprising another real-world data set focusing on loan applications, followed by Section~\nameref{sec:eval} where we assess the effectiveness of \textsc{VisRuler} by reporting the results of a user study. Subsequently, in Section~\nameref{sec:lim}, we discuss several limitations of our system and opportunities for future work. Finally, Section~\nameref{sec:con} concludes our paper.

\section{Related Work} \label{sec:relwo}
  According to a recent survey~\cite{Streeb2021Task} that has extensively analyzed tree- and rule-based classification, several VA systems have been developed for this topic in the \mbox{InfoVis} and VA communities. However, most of these tools do not employ algorithms and measures (except for the accuracy metric) in order to compare model quality.~\cite{Streeb2021Task} This section reviews prior work on the interpretation of bagged and boosted decision trees and the more general tools for tree- and rule-based visualization,
comparing them with \textsc{VisRuler} to highlight our tool's novelty.

\subsection{Interpretation of Bagged Decision Trees}
As in \textsc{VisRuler}, relevant works that utilize bagging methods use the RF algorithm to produce decision trees.~\cite{Zhao2019iForest,Neto2021Explainable,Eirich2022RfX,Nsch2019Colorful,Neto2021Multivariate} iForest~\cite{Zhao2019iForest}
provides users with tree-related information and an overview of the involved decision paths for case-based reasoning, with the goal of revealing the model's working internals. However, iForest can be used only for binary classification, while \textsc{VisRuler} can be used with multi-class data sets (as in the use case of Section~\nameref{sec:overview}). Also, the feature flow, a node-link diagram, suffers from scalability issues (a challenge only partially overcome with aggregation). Our tool employs dimensionality reduction for clustering all decisions extracted by multiple models, thus enabling users to gain insights into the patterns inside a large quantities of rules. Therefore, \textsc{VisRuler} allows users to mine rules for both a particular class outcome and in connection to a specific case. ExMatrix~\cite{Neto2021Explainable} is another VA tool for RF interpretation that operates using a matrix-like visual representation, facilitating the analysis of a model and connecting rules to classification results. While the scalability is good, it does not cover the task of finding similarities between decisions from diverse models and algorithms. In conclusion, none of the above works have experimented with the fusion of bagged and boosted decision trees, and in particular, with visualizing both tree types in a joint \emph{decisions space} to observe their dissimilarity, which can result in unique and undiscovered decisions. RfX~\cite{Eirich2022RfX} supports the comparison of several decision trees originating from a RF model with a dissimilarity projection and icicle plots, allowing electrical engineers to browse a single decision tree by using a node-link diagram. In contrast, \textsc{VisRuler} does not concentrate on a specific domain and gives attention to unique decision paths instead of trees with more scalable visual representations. Colorful trees~\cite{Nsch2019Colorful} follows a botanical metaphor and demonstrates many core parameters essential to comprehend how a RF model operates. This method allows customized mappings of RF components to visual attributes, thus enabling users to determine the performance, analyze the behavior of individual trees, and understand how to tune the hyperparameters to improve performance or efficiency. However, this work is targeted toward hyperparameter tuning and does not focus on concurrently extracting and analyzing the decisions from each RF and AB model. Additionally, it is impossible to accomplish case-based reasoning with the proposed visual representation. Finally, Neto and Paulovich~\cite{Neto2021Multivariate} describe the extraction and explanation of patterns in high-dimensional data sets from random decision trees, but model interpretation through the exploration of alternative decisions remains uncovered by this work (when compared to \textsc{VisRuler}).


\subsection{Interpretation of Boosted Decision Trees}
Special attention has been given to boosted decision trees with VA tools for diagnosing the training process of boosting methods~\cite{Liu2018Visual,Huang2019GBRTVis,Wang2021Investigating} and interpreting their decisions.~\cite{Xia2021GBMVis} Closer to our work, GBMVis~\cite{Xia2021GBMVis} aims to reveal the structure and properties of Gradient boosting,~\cite{Friedman2001Greedy} enabling users to examine the importance of features and follow the data flow for different decisions. A node-link diagram may limit its scalability to monitor hundreds or thousands of decisions concurrently, as opposed to \textsc{VisRuler}. Furthermore, our novel parallel coordinates plot adaptation allows users to instantly combine rules and observe their differences to identify unique decisions.
BOOSTVis~\cite{Liu2018Visual} employs views such as a temporal confusion matrix visualization for verifying the performance changes of the model, a t-SNE~\cite{vanDerMaaten2008Visualizing} projection for inspecting the instances, and a node-link diagram for examining the rules. Through GBRTVis,~\cite{Huang2019GBRTVis} users can explore Gradient boosting~\cite{Friedman2001Greedy} with a node-link diagram for the rules, the instances distribution shown in a treemap, and continuously monitoring the loss function. \emph{VIS}TB~\cite{Wang2021Investigating} contains a redesigned temporal confusion matrix to track the per-instance prediction during the training process. It also enables the comparison of the impact of individual features over iterations. These VA systems focus on the online training of boosting methods and aim to assist in feature selection and hyperparameter tuning. While these problems are (partially) tackled by our tool, we concentrate on interpreting the decisions from bagged and boosted decision trees and comparing them across models.

\subsection{Tree- and Rule-based Model Visualization}
Existing work on single decision tree visualization has experimented with different visualization techniques, such as node-link diagrams,~\cite{Elzen2011BaobabView,Nguyen2000A,Lee2016An,Cavallo2019Clustrophile,Barlow2001Case,Phillips2017FFTrees,Bremm2011Interactive,Hongzhi2004Multiple,Munzner2003TreeJuxtaposer,Behrisch2014Feedback} treemaps,~\cite{Muhlbacher2018TreePOD,Gomez2013Visualizing} icicle plots,~\cite{Padua2014Interactive,Ankerst2000Towards} star coordinates,~\cite{Teoh2003Starclass,Teoh2003PaintingClass} and 2D scatter plot matrices.~\cite{Do2007Towards} These techniques do not generalize well when exploring multiple decision trees, which is \textsc{VisRuler}'s primary design goal. Visualizing the surrogate models to approximate the behaviors of the original models, either globally or locally, is another branch of related works.~\cite{Yuan2022Visual,Cao2020DRIL,Castro2019Surrogate,Han2000RuleViz,Agus2021RISSAD,Ware2001Interactive,Yuan2021An,Eisemann2014A} Rule-based visualizations have also been deployed for the interpretation of complex neural networks.~\cite{Marcilio2021ExplorerTree,Ming2019RuleMatrix,Jia2020Visualizing,ThomasFacetRules} Nevertheless, these models differ due to the lack of inherent decisions that could be extracted directly from the bagged and boosted decision trees. The core mechanism of bagging and boosting methods is the generation of decisions based on the training data, which then experts can interpret.

Finally, multiple static visualizations and a few interactive VA tools have been developed for specific domains of research, such as medicine,~\cite{Hummelen2010Deep,Viros2008Improving,Li2020A,Niemman2014Learning,Carlson2008Phylogenetic} biology,~\cite{Abramov2019RuleVis,Sydow2014Structure} security,~\cite{Aupetit2016Visualization} and social sciences.~\cite{Moussaid2013Social} However, \textsc{VisRuler} is a model-agnostic solution that could be modified to work with various domains, depending on the given data set and the domain expert. 
%
%

\subsection{Visual Analytics for Multi-model Comparison}
Several VA systems exist that enable the comparison of ML models in classification problems, especially with the evaluation of predictive performance and the importance of features.~\cite{Xu2019EnsembleLens,Schneider2018Integrating,Talbot2009EnsembleMatrix,Zhang2019Manifold,Squares2017Ren,Gleicher2020Boxer,Das2020QUESTO,Li202020A,Ono2021Pipeline,Chatzimparmpas2021StackGenVis,Chatzimparmpas2021VisEvol} EnsembleLens~\cite{Xu2019EnsembleLens} is a VA system working with multiple models trained into different feature subsets. The end goal is to visualize the correlation between ensemble models, algorithms, hyperparameters, and features that achieve the highest score for anomalous cases. On the contrary, \textsc{VisRuler} is not limited to anomaly detection problems, instead focusing on the interpretability of insightful rules extracted from two ensemble algorithms that produce tree-based decisions. Schneider et al.~\cite{Schneider2018Integrating} explored the impact of comparing side-by-side the data and model spaces. They used both bagging and boosting ensembles and investigated these algorithms' influence on the data with the purpose of adding, deleting, or replacing models from the model space. We also support that this mixture of bagging and boosting models is beneficial because each algorithm can bring different information to the analysis of decision rules; for more details, please check Section~\nameref{sec:back}. However, we abstract complex models into individual decisions that are accountable to and easily interpretable by users.~\cite{Breiman2001Statistical}

EnsembleMatrix~\cite{Talbot2009EnsembleMatrix} and Manifold~\cite{Zhang2019Manifold} are two VA tools specifically designed for model comparison. The former uses a confusion matrix representation for contrasting models. The latter produces and compares pairs of models across all data classes. We adopt a similar approach as with those tools, but instead of deciding which model was ``optimal'', we export all decisions that work well with a specific test instance to be examined by domain experts. Squares~\cite{Squares2017Ren} is a visualization approach that showcases the per-class performance of a multi-class data set. It helps users prioritize their efforts by calculating common validation metrics. Similarly, Boxer~\cite{Gleicher2020Boxer} is a system that faciliates the interactive exploration of subsets of training and testing data while comparing the performance of multiple models on those instances. Another VA system, called QUESTO,~\cite{Das2020QUESTO} enables domain experts to control objective functions to set specific constraints and to search for an ``optimal'' model for this purpose. Li et al.~\cite{Li202020A} developed a VA system that allows multi-model comparison based on clinical data predictions with a special attention to feature contribution. \emph{PipelineProfiler}~\cite{Ono2021Pipeline} is a visualization tool for the exploration of several AutoML~\cite{automl} pipelines comprising multiple models. StackGenVis~\cite{Chatzimparmpas2021StackGenVis} is a VA system for composing powerful and diverse stacking ensembles~\cite{Wolpert1992Stacked} from a pool of base models. Finally, VisEvol~\cite{Chatzimparmpas2021VisEvol} is a VA tool that supports the interactive intervention in the evolutionary hyperparameter optimization process while exploring five alternative ML algorithms. On the one hand, we also facilitate users to assess various models. On the other hand, we focus on making the process of model-learned decision rules more transparent and justifiable with the involvement of a domain expert at specific phases (see Figure~\ref{fig:collaboration-diagram} and Section~\nameref{sec:overview}).

There is also a body of literature devoted to regression problems.~\cite{Muhlbacher2013APartition,Sehgal2018Visual,Zhao2014LoVis,Das2019BEAMES} For instance, BEAMES~\cite{Das2019BEAMES} contains four ML algorithms and a model sampling method, with the help of which the system generates an ordered list of models that aids the user in selecting a performant model. Although feature ranking is also covered by this tool, ML experts working with \textsc{VisRuler} aim at gathering several manual decisions extracted from two directly comparable algorithms that the domain expert should interpret. Finally, classification requires different handling in terms of the available algorithms and validation metrics when compared to regression tasks, making such solutions challenging to adapt to classification problems and vice versa.

\section{Random Forest vs. Adaptive Boosting} \label{sec:back}
  In this section we present a quick overview of the general algorithmic steps of (and differences between) the RF and AB algorithms, in order to familiarize potential readers with the techniques involved and to highlight the importance of our tool. For more details, please refer to the extensive literature published on the two algorithms since their seminal works.~\cite{Freund1999A,Breiman2001Random} 

Suppose there are $N$ instances in our training data set. If we consider a binary classification problem, then we will have: ${x}_i \in \mathbb{R}^n$, ${y}_i \in \{-1,1\}$, where $n$ is the number of features, $x$ is the set of instances with $i = 1, 2, \dots, N$, and $y$ is the target variable which is either $-1$ or $1$, 
designating either class ${C}1$ or ${C}2$.

\textbf{Random Forest.} This algorithm works in two stages. The first stage involves integrating numerous decision trees to construct the RF, and the second stage involves making predictions for each tree created in the first stage, followed by a majority voting strategy.

The algorithm starts by looping for $m = 1$ to $M$, where $M$ is the \emph{number of trees/estimators} hyperparameter. Then, a bootstrap sample $\mathbb{Z}^*$ of size $N$ is drawn from the training data by sampling $N$ times with replacement. Afterwards, a decision tree $T_m$ is grown with the bootstrapped data by recursively repeating the following steps for each terminal node of the tree, until specific conditions have been met (e.g., the \emph{maximum depth} of a tree $d_{\mathrm{max}}$ is achieved). The first step is to form individual decision trees: $n$ features, their number is limited by the \emph{maximum number of features} hyperparameter, and it is selected at random from the $N$ instances (defined in the previous paragraph). Next, the best \emph{split point} $p_v$ among the $n$ is picked. Finally, the node $v$ is split into two child nodes.

The output is an ensemble of trees $\{T_m\}_1^M$ that can be used to make predictions at a new test instance $k$, as follows. Let $\hat{C}_m(k)$ be the class prediction of the $m$th decision tree; the output will be calculated as: $\text{majority\_vote}\{ \hat{C}_m(k)_1^M \}$. For a test instance, the \emph{decision path} that predicts it is followed repeatedly for every decision tree, and the test instance gets assigned to the category that wins the majority votes.

\begin{figure*}[tb]
\centering
\includegraphics[width=\linewidth]{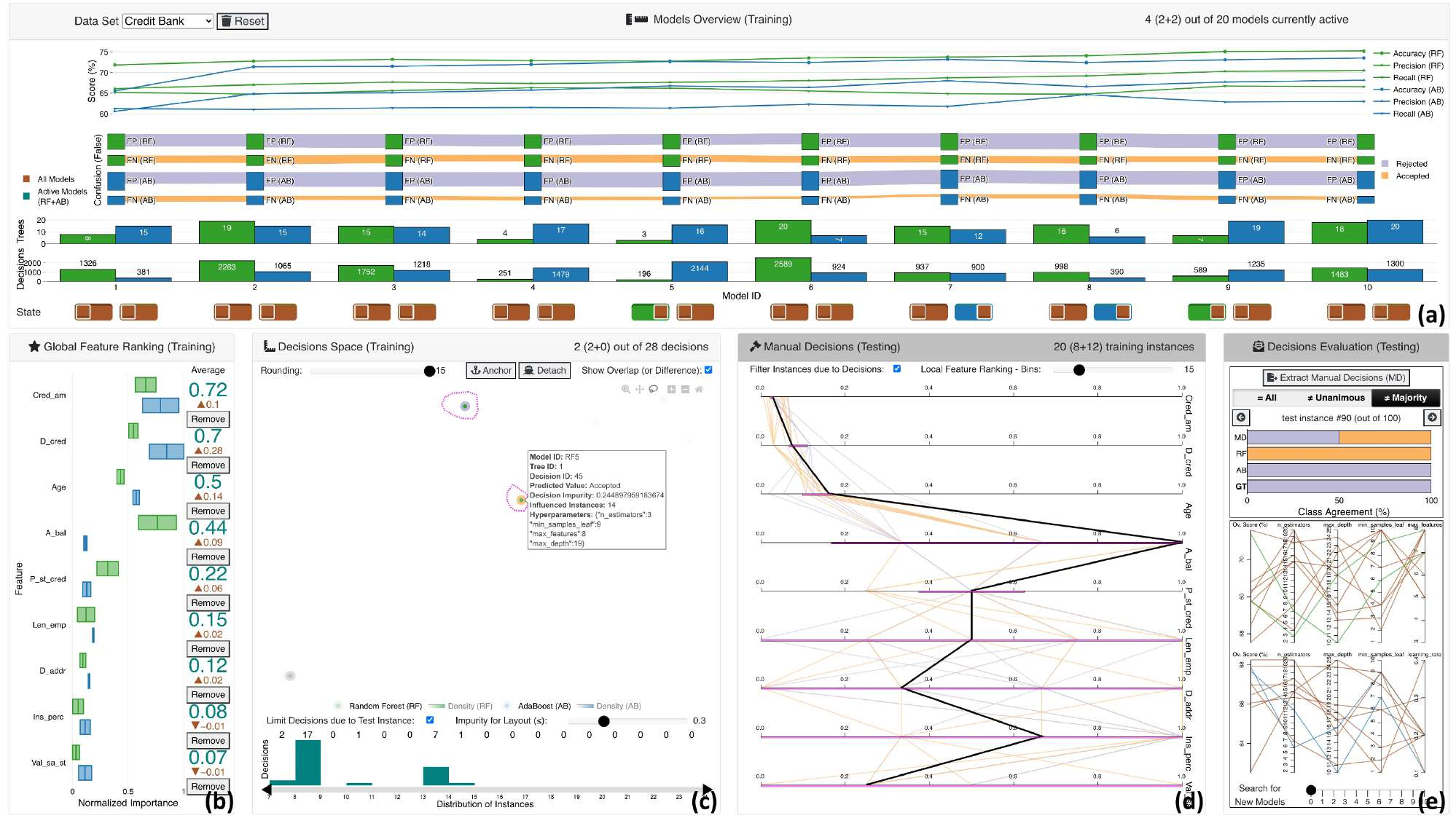}
 \caption{Extracting decision rules for manual evaluation with \textsc{VisRuler}: (a) panel with visual metaphors for selecting performant and diverse models; (b) box plot for feature selection according to per algorithmic importance; (c) visual embedding of computed decisions that training instances fall in due to their values; (d) vertical parallel coordinates plot that summarizes the rules with value ranges for each feature and highlights the current test instance; and (e) horizontal stacked bar chart for revealing the class agreement of each model against the manual decisions, together with the parallel coordinates plots for tuning hyperparameters and training new models.}
 \label{fig:teaser}
\end{figure*}

For each $v$ in a decision tree, the $p_v$ refers to a feature and different criteria (e.g., the \emph{minimum samples} in each leaf of a tree $s_{\mathrm{min}}$ is reached) tunable through the hyperparameter settings that are used to split this node to children nodes. The \textbf{Gini impurity}~\cite{Genuer2010Variable} measurement is utilized in our case to pick this $p_v$, formulated for each $v$ as follows: $GI(v) = \sum_{c = 1}^C P_{vc} \cdot (1 - P_{vc})$, where $P_{vc}$ is the probability of a certain classification $c$ (in our example ${C}1$ and ${C}2$). Given an input test instance, each decision tree will fall into a root-to-leaf decision path, which leads to its prediction. Every \textbf{decision path} (or simply \emph{decision}) is constructed by defining a range of minimum and maximum values for each feature extracted from RF and AB grown decision trees. Therefore, for $n$ features, we will have $n \cdot 2$ ranges, i.e., dimensions for projecting the decisions visible in Figure~\ref{fig:teaser}(c).

\textbf{Adaptive Boosting.} This algorithm fits successively many decision stumps or even decision trees to the training data, using various weights. The latter occurs only if the \emph{maximum depth} hyperparameter is set to $>1$
. It begins by forecasting the original data set and weighting each observation equally. If the first decision stump's prediction is inaccurate, the observation that was mistakenly anticipated is given more weight. Because it is an iterative process with a specific improvement step controlled by the \emph{learning rate} hyperparameter, it will continue to add decision stumps until the \emph{number of trees/estimators} reaches a limit.

For each instance, AB calculates the weights. Each training example is given a weight to determine its importance in the training data set. When the given weights are substantial, that set of training instances is more likely to influence the training set and vice versa. All training instances will start with the same weight, defined as: $w_i = 1/N$. The weighted samples always add up to one, and each individual weight's value will be between 0 and 1.

The algorithm starts by looping for $m = 1$ to $M$, fitting a model $G_m(x)$ to the training data using weights $w_i$. After that, AB uses the method to calculate the real effect of this classifier in categorizing the training instances: $\alpha_m = \log((1 - \text{err}_m) / \text{err}_m)$, where $\text{err}_m$ is the total number of misclassifications for that training set divided by the training set size. Therefore, $\alpha_m$ represents how influential this stump will be in the final categorization. When a decision stump performs well or has no misclassifications, the error rate is 0 and the $\alpha$ value is relatively significant and positive. The alpha value will be 0 if the stump only classifies half correctly and half erroneously. Finally, the $\alpha$ would have a big negative number if the stump consistently produced misclassified data.

After plugging in the actual values of total error for each stump, the sample weights are updated. The following formula is used to accomplish that: $w_i = w_{i-1} \cdot \euler^{\alpha}$.
In detail, the new sample weight will be equal to the old sample weight multiplied by Euler's number, raised to $\alpha$. As explained in the previous paragraph, the two cases for $\alpha$ are: (a) positive when the predicted and the actual output agree or (b) negative when the predicted output does not agree with the actual class (i.e., the sample is misclassified). In the first case, the sample weight is decreased from what it was before, since the algorithm already performs well. In the second case, the sample weight gets increased so that the same misclassification does not repeat in the next stump. This procedure is followed so that the stumps are dependent on their predecessors. Lastly, the output for AB is the sum of all trees: $G(x) = \pm \bigl[ \sum_{m = 1}^M \alpha_m G_m(x) \bigr]$.

\textbf{Differences.} There are two key distinctions between bagging and boosting, which apply for the RF and AB algorithms. First of all, boosting changes the distribution of the training set adaptively based on the performance of previously constructed classifiers, whereas bagging changes the distribution of the training set stochastically. Second, boosting employs a function of a classifier's performance as a weight for voting, whereas bagging uses equal weight voting. Boosting appears to minimize both bias and variance, unlike bagging, which is primarily for variance reduction. Following the training of a decision stump, the weights of misclassified training examples are raised while the weights of correctly classified training examples are dropped in order to train the following decision stump. Thus, efforts to overcome the bias of the most recently generated weak model by concentrating greater attention on the samples that it misclassified gain traction. Because of its capacity to minimize bias, boosting works particularly effectively with high-bias but low-variance weak models. The fundamental issue with boosting appears to be resilience to noise.~\cite{Bauer1999Empirical} This is to be expected, as noisy cases are more likely to be misclassified, and their weight will rise as a result.

To sum up, for data with little noise, boosting is regarded as being stronger than bagging; nevertheless, bagging algorithms are far more resilient than boosting in noisy environments.~\cite{Kotsiantis2007Combining} To mitigate this problem, using both algorithms to derive decisions is a novel idea we adopt. In the future, we may expand our approach to experiment with any decision-based algorithm (\hl{as in prior works}~\cite{Kotsiantis2007CombiningBagging,Kotsiantis2011Combining}), but in this paper, we chose to combine two already widely-used algorithms (i.e., RF and AB).

\section{Target Groups, Design Goals, and Analytical Tasks} \label{sec:goals}

In this section, we specify the main design goals (\textbf{G1--G5}) upon which \textsc{VisRuler} (or any other VA system) should base its development regarding scenarios of extracting decision rules from bagging and boosting ensemble algorithms. Afterwards, we report the corresponding analytical tasks (\textbf{T1--T5}) that a user should be capable of performing while he/she obtains assistance and guidance from our proposed system. One important aspect of the design of \textsc{VisRuler} is the need for collaboration between different experts, which we also motivate here.


\subsection{Target Groups}

In the InfoVis/VA communities, most of the research in explainable ML focuses on assisting \emph{ML experts and developers} in understanding, debugging, refining, and comparing ML models.~\cite{Chatzimparmpas2020A,Chatzimparmpas2020The} In this paper, we expand our method to involve another target group: the various \emph{domain experts} affected by the ML progress in fields such as finance, social care, and health care. With the growing adoption of ML in different areas, domain experts with little knowledge of ML algorithms might still want (or be required) to use them to assist in their decision-making. On the one hand, their trust in such decisions could be low due to a lack of in-depth knowledge on how models are learning from the training data. On the other hand, ML experts often have little prior knowledge about the data from particular domains. Thus, the primary goal of \textsc{VisRuler} is to combine the best of both worlds, i.e., to offer a solution that combines the above-mentioned benefits from both expert groups. More details about the collaboration between the ML and domain experts can be found in Section~\nameref{sec:overview}.

\subsection{Design Goals}

Our design goals originate from the analysis of the related work in Section~\nameref{sec:relwo}, especially the three design goals from Zhao et al.~\cite{Zhao2019iForest} (\textbf{G2 \& G3}) and the four questions from Ming et al.~\cite{Ming2019RuleMatrix} (\textbf{G1, G4, \& G5}) targeted to experts in domains such as health care, finance, security, and policymakers. 
Our methodology is similar to Zhao et al.,~\cite{Zhao2019iForest} who reviewed 35 papers from the ML, visualization, and human-computer interaction (HCI) communities to come up with their decision goals.

\textbf{G1: Bring diverging models' performance to the spotlight.}
A VA system must first focus on what each model has learned in general; as such, the assessment of every model's performance (to decide which should remain under use) is a prerequisite. Models that fail to perform according to user-defined standards should not be part of the following procedure.

\textbf{G2: Disclose connections between features and predictions.}
VA systems should expose the features' impacts on predictions and allow humans to delete needless features based on that. During training, ML models learn different mappings between input features and resulting predictions based on the inherent mathematical functions used and the setting of hyperparameters. These mappings describe model behavior and help humans comprehend RF and AB models' properties.

\textbf{G3: Discover the core hidden operating processes.}
The underlying functioning processes of RF and AB models must also be revealed to check whether the models are working correctly and to understand why a given prediction was made. Before making a decision, humans should be able to audit the decision process of a prediction and ensure that they agree. Many RF and AB model interpretation problems can be resolved by analyzing the individual decision paths.

\textbf{G4: Reason about the relationship of certain features and knowledge acquired.}
Unlike \textbf{G2}, this one concentrates on deviations in ranking features for justifying the rule-based generated knowledge. VA systems can support humans with this alignment of features and knowledge extracted through the decision rules. Since domain experts may have knowledge and ideas based on years of research and study that current ML models do not make use of, the facilitation of communication between ML models and humans is a primary design goal.

\textbf{G5: Guide to unconfident and contradictory predictions.} 
This goal emerges when a model fails to perform well on particular test instances. In the production deployment of ML models, a rule that some models are confident about may not be generalizable. Although undesirable, it is relatively common for models to produce contradictory predictions for certain difficult-to-classify test instances. As a result, VA systems must advise users on which occasions each model failed to predict correctly.

\subsection{Analytical Tasks}
To fulfill our design goals, we have determined five analytical tasks that should be supported by our VA system (described in Section~\nameref{sec:overview}).

\textbf{T1: Compare the performance and architecture of models for selecting the most effective ones.}
Users should be able to compare different models with the support from various measurements (\textbf{G1}), as follows: (1) illustrate the performance of each model based on multiple validation metrics; (2) distill the number of false-positive and false-negative instances from the confusion matrix for every model; and (3) derive the number of decision trees and decision paths per model, to facilitate high-level comparison.

\textbf{T2: Investigate the contribution of global features according to different models and algorithms.}
Following the preceding task, users should be guided through the process of selecting important features  (\textbf{G2}). Thus, it is crucial to enable the comparison between per-algorithm and per-model feature importances.

\textbf{T3: Explore alternative clusters of decisions for global explanation and case-based reasoning.} The summarization of the decisions in a single view that combines the decisions of different algorithms and models should be accomplished to allow users to assess the influence of each decision (\textbf{G3}). For example, some decisions could overfit, and others could contain a mixture of instances falling in different classes. This last phenomenon increases their impurity. Users should be able to interact and explore this \emph{decisions space}.

\textbf{T4: Compare decision rules based on local feature ranking.} The global features described in~\textbf{T2} might not be similarly important for specific decisions, hence, local feature ranking via contrastive analysis~\cite{Zou2013Contrastive} could shed some light upon this task (\textbf{G4}). Moreover, the interpretation of rules extracted from the space of solutions (see~\textbf{T3}) could be achieved if users are capable of investigating the values of both training and testing instances.

\textbf{T5: Identify the different types of failure cases and confrontation via manual decisions.} Failure to converge to a certain result due to the disagreement of the ML models should be highlighted to users (\textbf{G5}). For instance, if there is no uniformity in the final decision or the majority voted for the wrong result, it could be that these instances are outliers, borderline cases, or simply misclassified; being able to explore such cases is essential.

\section{System Overview and Use Case} \label{sec:overview}
Motivated by the above design goals and tasks, we have developed \textsc{VisRuler}. Our VA tool is written in Python and JavaScript. More technical details available on GitHub.~\cite{VisRulerCode}

\begin{figure}[tb]
\centering
\includegraphics[width=\columnwidth]{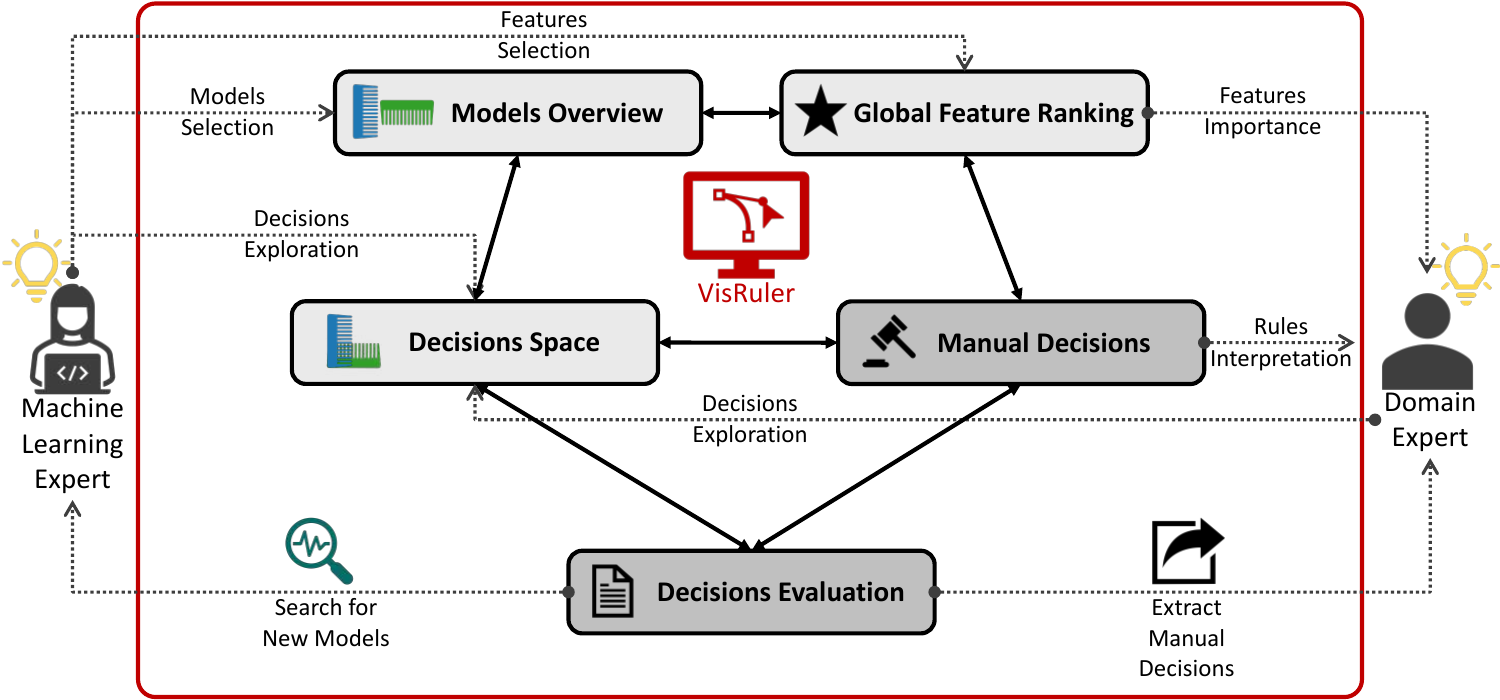}
\caption{The \textsc{VisRuler} workflow allows ML experts to select performant and diverse models, choose important features, and retrain models with new hyperparameters. Domain experts can explore robust decisions, compare them to global standards, identify local decisions for a specific test instance, and extract them.}
\label{fig:workflow-diagram}
\end{figure}

The tool consists of five main interactive visualization panels (Figure~\ref{fig:teaser}): (a) \emph{models overview} (\textbf{T1}), (b) \emph{global feature ranking} (\textbf{T2}), (c) \emph{decisions space} (\textbf{T3}), (d) \emph{manual decisions} (\textbf{T4}), and (e) \emph{decisions evaluation} (\textbf{T5}).
Our proposed \textbf{workflow} is a two-party system with the ML expert on the one side and the domain expert on the other (see Figure~\ref{fig:workflow-diagram}). The above-mentioned panels of our tool support the experts' collaborative effort, specifically: (i) the ML expert should select powerful and diverse models from the two separate algorithms based on their performance assessed by validation metrics (Figure~\ref{fig:teaser}(a));
(ii) during this phase, the ML expert should choose which features are important for the active models compared to all models (see Figure~\ref{fig:teaser}(b));
(iii) in the next exploration phase, both experts should examine which decisions explain the data set globally and decide upon impactful decisions for a specific test instance (cf. Figure~\ref{fig:teaser}(c)); 
(iv) in this same phase, the domain expert should interpret the manual decisions selected in order to gain insights about the models' decisions---either globally or locally---for a particular test instance (Figure~\ref{fig:teaser}(d)); and 
(v) in the final phase, the domain expert can evaluate the agreement and extract suitable manual decisions while the ML expert should search for new models if the search did not reach a satisfactory level according to the domain expert (Figure~\ref{fig:teaser}(e)). Overall, this is an iterative process with a final goal to receive insightful decisions that should be interpretable for all counterparts. Details about the different views within the panels can be found below.

\begin{figure}[t]
\centering
\includegraphics[width=\columnwidth]{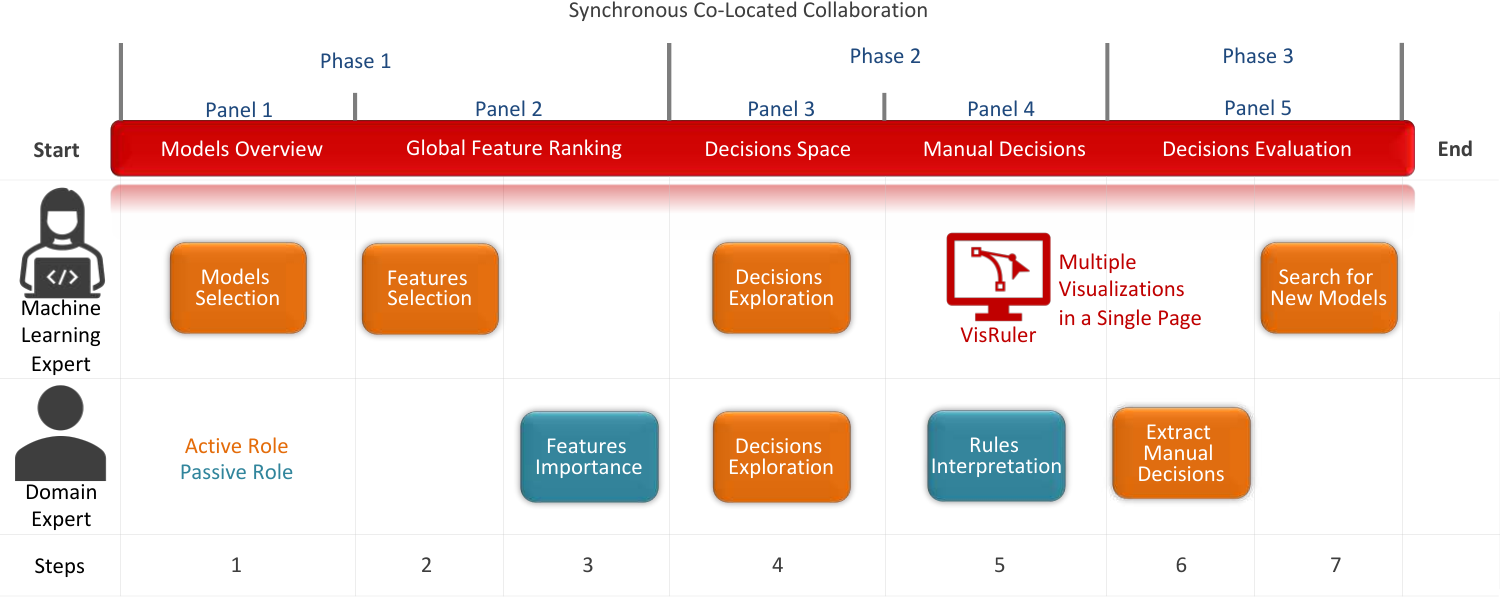}
\caption{The \textsc{VisRuler} cooperation diagram illustrates how synchronous co-located collaboration typically happens between the ML expert and the domain expert. Three phases and five panels support their teamwork in a single-page tool, with the ML expert being more active (orange color) than the domain expert who receives and analyzes information (teal color). The seven linear steps taken by each user are also noted at the bottom.}
\label{fig:collaboration-diagram}
\end{figure}

\textsc{VisRuler} incorporates a single workflow for the synchronous co-located collaboration between the ML expert and the domain expert, as depicted in Figure~\ref{fig:collaboration-diagram}. It is a VA system that comprises multiple coordinate views arranged in a single webpage to manage the entire process without any distractions occuring due to the navigation to different tabs. The three phases and five visualization panels are already described in the previous paragraph. Both experts' usual interactions with the system can be aggregated into seven steps that are followed in the use case explained in this section and the usage scenario in Section~\nameref{sec:case}. Most of the time, the active role (cf. orange color in the diagram) is given to the ML expert who is responsible for selecting models (step 1), selecting features (step 2), exploring decisions (step 4), and finally, in step 7, search for new models (ideally based on the domain expert's feedback). On the other hand, the domain expert often has a more passive role (see teal color in the diagram) since he/she should focus on important features (step 3) and interpret the decision rules (step 5) based on the prior step. But in step 6, the domain expert becomes active because he/she has to decide which manual decisions should be extracted, for example, to be used as input to another tool, or as evidence for his/her diagnosis. The meeting point between both experts is step 4, the exploration of decisions, where the ML expert can control the multiple implemented filtering options (e.g., acceptable impurity level) while the domain expert pinpoints specific decision paths that could be interesting for a detailed manual investigation. \hl{Without their continuous communication while they operate the system and observe each other's actions, gaining deep insights would be very difficult.}

The workflow of \textsc{VisRuler} is model-agnostic, as long as rules can be extracted from the ML algorithms. 
Currently, the implementation uses two popular EL methods: (1) RF and (2) AB (cf. green and blue colors in Figure~\ref{fig:workflow-diagram}, respectively). 
This choice was intentional since bagging methods work differently than boosting, as explained in Section~\nameref{sec:back}. 
Furthermore, each data set is split in a stratified fashion (i.e., keeping the class balance in training/testing split) into 90\% of training samples and 10\% of testing samples. We also use cross-validation with 3-folds on the training set, and we scan the hyperparameter space for 10 iterations using Random search~\cite{Bergstra2012Random} in each algorithm separately. The common hyperparameters for both ML algorithms we experimented with (and their intervals) are: number of trees/estimators (2--20), maximum depth of a tree (10--25), and minimum samples in each leaf of a tree (1--10). 
An extra hyperparameter of RF is the maximum number of features to consider when looking for the best split (($\sqrt{number\_of\_features}$)--($number\_of\_features-1$)). AB has the learning rate (0.1--0.4). It is straightforward to adapt those values through the code.

In the following subsections, we explain \textsc{VisRuler} by describing a use case with the \emph{World Happiness Report 2019}~\cite{Helliwell2019World} data set obtained from the Kaggle repository.~\cite{Kaggle2019} This data set contains 156 countries (i.e., instances) ranked according to an index representing how happy the citizens of each country are. The six other variables that could be considered as features are: (1) \emph{GDP per capita}, (2) \emph{social support}, (3) \emph{healthy life expectancy}, (4) \emph{freedom to make life choices}, (5) \emph{generosity}, and (6) \emph{corruption perception}. Because this data set does not contain any categorical class labels, we follow the same approach as in Neto and Paulovich~\cite{Neto2021Multivariate} to discretize the happiness score in three different bins. Hence, we are converting this regression problem into a multi-class classification problem.~\cite{Salman2012Regression} Also in our case, the original variable Score becomes the target variable that our ML models should predict. In detail, the HS-Level-3 class contains 42 countries with happiness scores (HS) ranging from $6.13$ to $7.76$, the HS-Level-2 groups 79 countries from $4.49$ to $6.13$, and the HS-Level-1 class encloses 35 countries from $2.85$ to $4.49$.

\subsection{Models Overview} \label{sec:models}

The exploration starts with an overview of how 10 RF and 10 AB models performed based on three validation metrics: accuracy, precision, and recall. The models are initially sorted according to the overall score, which is the average sum of the three metrics. \hl{This choice guides users to focus mostly on the right-hand side of the line chart (as showcased in Section~\nameref{sec:application}).} Green is used for the RF algorithm, while blue is for AB. All visual representations share the same x-axis: the identification (ID) number of each model. \hl{The design decision to align views vertically enables us to avoid repetition and follows the best practices.} The line chart in Figure~\ref{fig:teaser}(a) always presents the worst to best models from left to right. The y-axis denotes the score for each metric as a percentage, with distinct symbols used for the different metrics. 
The confusion plot inspired by Sankey diagrams in Figure~\ref{fig:teaser}(a) visually maps a confusion matrix of only false-positive and false-negative values for each model into nodes with different heights depending on the number of confused training instances. Then, they are divided into two groups reflecting the two algorithms. It also presents the confusion of all individual classes for the different instances when comparing two subsequent models, as illustrated in both Figure~\ref{fig:teaser}(a) and Figure~\ref{fig:use_case1_model}(a). The width of the band between two consecutive nodes indicates the increase or decrease in confusion from one model to the other sequentially, so the smaller the height of a line, the better a model's prediction compared to its predecessor or successor. The same effect applies to each node that absorbs the lines. With this plot, users can focus on the misclassified training instances that are more important for a given problem. For example, a medical doctor is typically cautious when dealing with false-negative instances since human lives may be at risk. Users can also check how many misclassified instances exist in each model and propagate from one model to another for each label class. \hl{Inspired by previous works,~\cite{Liu2018Visual,Wang2021Investigating} we utilize a distinct visual metaphor for this plot to convey---as concisely as possible---the per class confusion for the several under examination ML models.}
The bar charts in Figure~\ref{fig:teaser}(a) showcase the two main architectural components of the bagged and boosted decisions trees, which are the \emph{number of trees/estimators} hyperparameter and the number of decisions generated from these trees for every model mapped in the y-axes, respectively. These visualizations allow users to check the related hyperparameters of the individual models in a juxtaposed manner, since the number of decisions is related to the number of trees and the maximum allowed depth of each tree (i.e., \emph{max\_depth} hyperparameter). Finally, the state shown in Figure~\ref{fig:teaser}(a) designates which models are currently active (green or blue, respectively). In order to enable the comparison between the currently active model against all models, each icon for an active model contains a brown-colored slider thumb (Figure~\ref{fig:teaser}(a), legend on the left).

\subsection{Global Feature Ranking} \label{sec:features}

The box plots which aggregate per-algorithm importance (see~Figure~\ref{fig:teaser}(b)) provide a holistic view of the performance of the models. Each pair of boxes is related to a unique feature, summarizing the active models' normalized importance per feature (from 0 to 1, i.e., worst to best). The box plots are sorted according to the average values of all active models, visible as a number in teal. The difference to all models being active is shown with arrows facing up for increase or down for decrease in per-feature importance. For both algorithms, we compute feature importance as the mean and standard deviation of accumulation of the impurity decrease within each tree. This is a measurement that can be calculated directly from RF~\cite{Rogers2006Identifying} and AB~\cite{Wang2012AdaBoost} algorithms, cf. Section~\nameref{sec:back}.

\subsection{Decisions Space} \label{sec:space}

\begin{figure*}[tb]
\centering
\includegraphics[width=\linewidth]{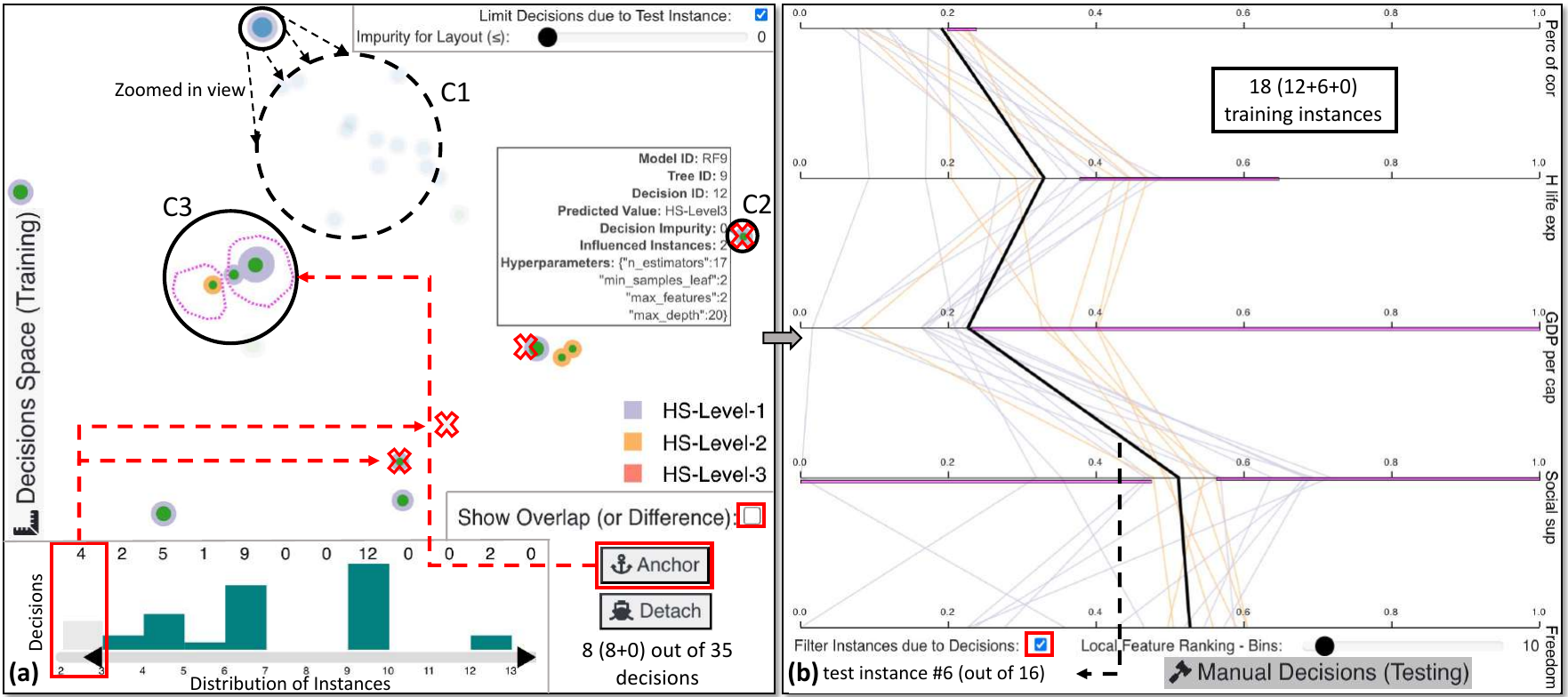}
\caption{The local exploration of the \nth{6} test instance with specific decisions from all the active models that apply only for this and similar test instances. (a) is a projection-based view that includes \circled{C1} with multiple impure decisions, visible only after zooming in. \circled{C2} is a decision with only 2 influenced training instances falling in this path, hence, we interact with the histogram below it to filter out overfitting cases with only 2 or 3 training instances. The comparison mode is enabled for \circled{C3}, resulting in anchoring two subclusters of different (in terms of class) prediction decisions. In (b), the PCP highlights the differences between these previous subclusters for each feature in magenta color. This view is dynamic since the features are constantly being re-sorted based on contrastive analysis of the selected cluster against all points.}
\label{fig:use_case1_broderline}
\end{figure*}

The projection-based view in Figure~\ref{fig:teaser}(c) is produced with the UMAP algorithm,~\cite{McInnes2018UMAP} selected due to its popularity and the results of a recent quantitative survey~\cite{Espadoto2021Toward} that found this algorithm the best overall among many others. In the visual embedding, decision paths are clustered based on their similarity according to their ranges for each feature, as described in Section~\nameref{sec:back} and in the work of Zhao et al.~\cite{Zhao2019iForest} Therefore, it enables an algorithm-agnostic comparison between decisions stemming from both RF and AB models.
The green color in the center of a point indicates that a decision is from RF, while blue is for AB. The outline color reflects the training instances' class based on a decision's prediction. The size maps the number of training instances that are classified by a specific decision, and the opacity encodes the impurity of each decision. Low impurity (with only a few training instances from other classes) makes the points more opaque. The positioning of the points can be used to observe if the RF and AB models produced similar rules, offering a comparison between algorithm decisions. The histogram in Figure~\ref{fig:teaser}(c) shows the number of decisions (y-axis) and the distribution of training instances in these paths (x-axis), and can also be used to filter the number of visible decisions in the projection-based view to avoid overfitting rules containing only a few instances (as shown in Figure~\ref{fig:use_case1_broderline}(a)) or general rules that might not apply in problematic cases.

UMAP is initiated with variable \emph{n\_neighbors} and \emph{min\_dist} fixed to $0.1$. To determine the optimal number of clusters to be visualized, DBSCAN~\cite{Ester1996A} is used to compute an estimated number of core clusters from the derived decisions, which is then used to tune the \emph{n\_neighbors}, with a minimum of 2 and a maximum of 100 neighbors (the aim is to have the same magnitude in both). For the first experiment in Section~\nameref{sec:application}, \emph{n\_neighbors} was automatically set to 20. On the other hand, in the usage scenario of Section~\nameref{sec:case}, DBSCAN estimated 477 clusters, which tuned the hyperparameter to the maximum value.

Multiple interactions are possible in this entire panel. The rounding slider (set to 15) allows users to round all decisions' range values to the desired decimal points. The comparison mode (active in Figure~\ref{fig:teaser}(c)) enables users to anchor groups of points and compare the selection against any other cluster. The two alternative choices are to present either the overlap or difference between the handpicked groups (shown in magenta color); the \emph{Detach} button is for canceling this mode. Density views assist users in observing the distribution of RF against AB decisions in the projection, which is helpful if large amounts of decisions are visualized, as illustrated in Figure~\ref{fig:use_case1_safe}(a), projection. The \emph{Limit Decisions due to Test Instance} checkbox alters the layout and changes global decisions' exploration to local for a particular case. Finally, a limit can be set for the acceptable impurity that is visible. If a decision is more impure than the currently chosen value, then it becomes almost transparent. As this view is tightly connected with the visualization of the following view, we proceed directly to Section~\nameref{sec:manual}.

\begin{figure*}[tb]
\centering
\includegraphics[width=\linewidth]{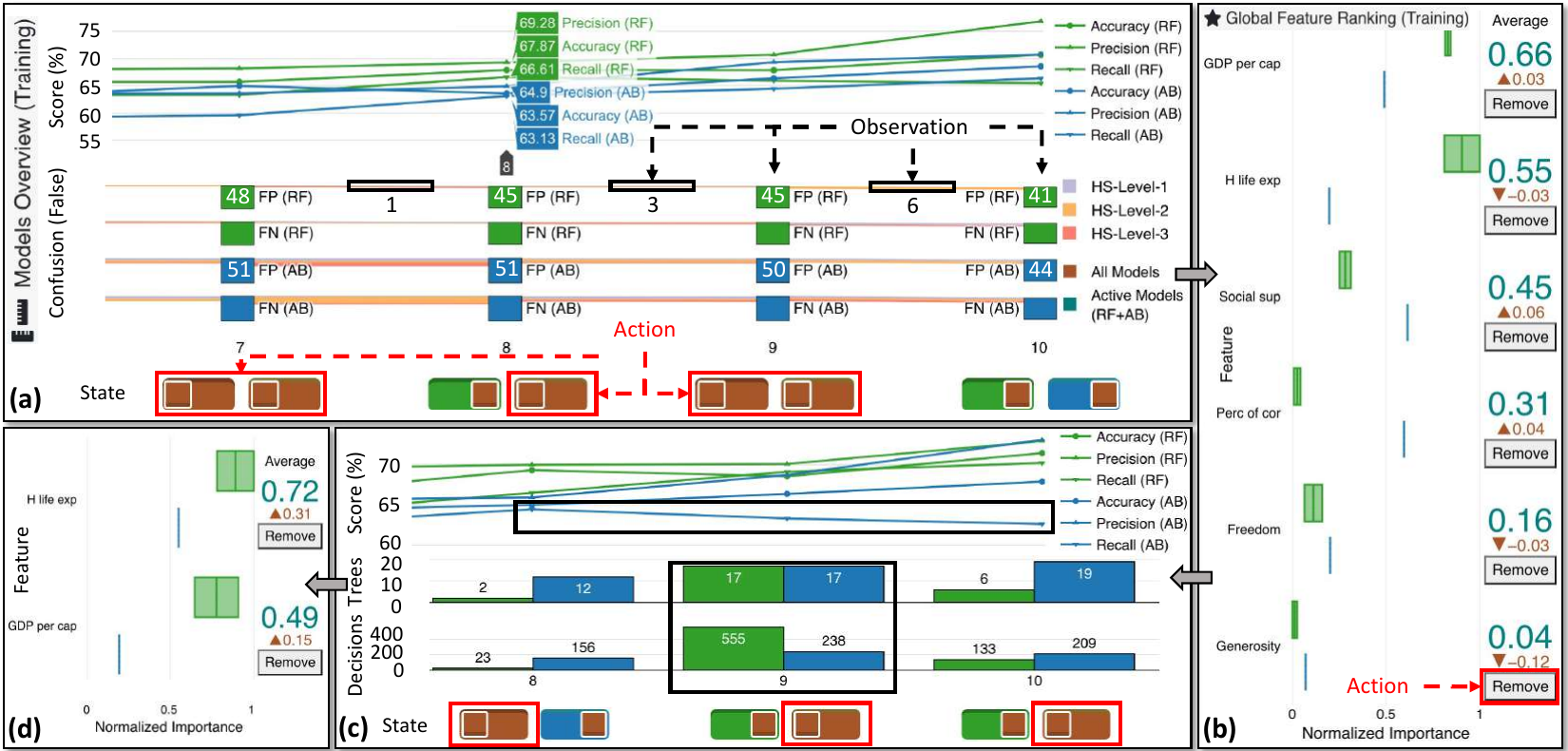}
\caption{Exploration of ML models with \textsc{VisRuler}. View (a) presents the deactivation of all models except for RF8, RF10, and AB10, after consideration of their performance based on multiple metrics displayed in the visualizations. In (b), \emph{Generosity} is the least important feature for the three active ML models and, particularly, its importance decreased while we deactivated most of the available ML models (see brown color). View (c) indicates that, after retraining with 5 of 6 original features, the new AB8 is better than the subsequent models due to the decline in recall; AB8, RF9, and RF10 remain the only active models after this step. In the box plot (d), the feature \emph{H life exp} becomes more important by far than \emph{GDP per cap}. Thus, these features swapped places compared to view (b).}
\label{fig:use_case1_model}
\end{figure*}

\subsection{Manual Decisions} \label{sec:manual}

\begin{figure*}[tb]
\centering
\includegraphics[width=\linewidth]{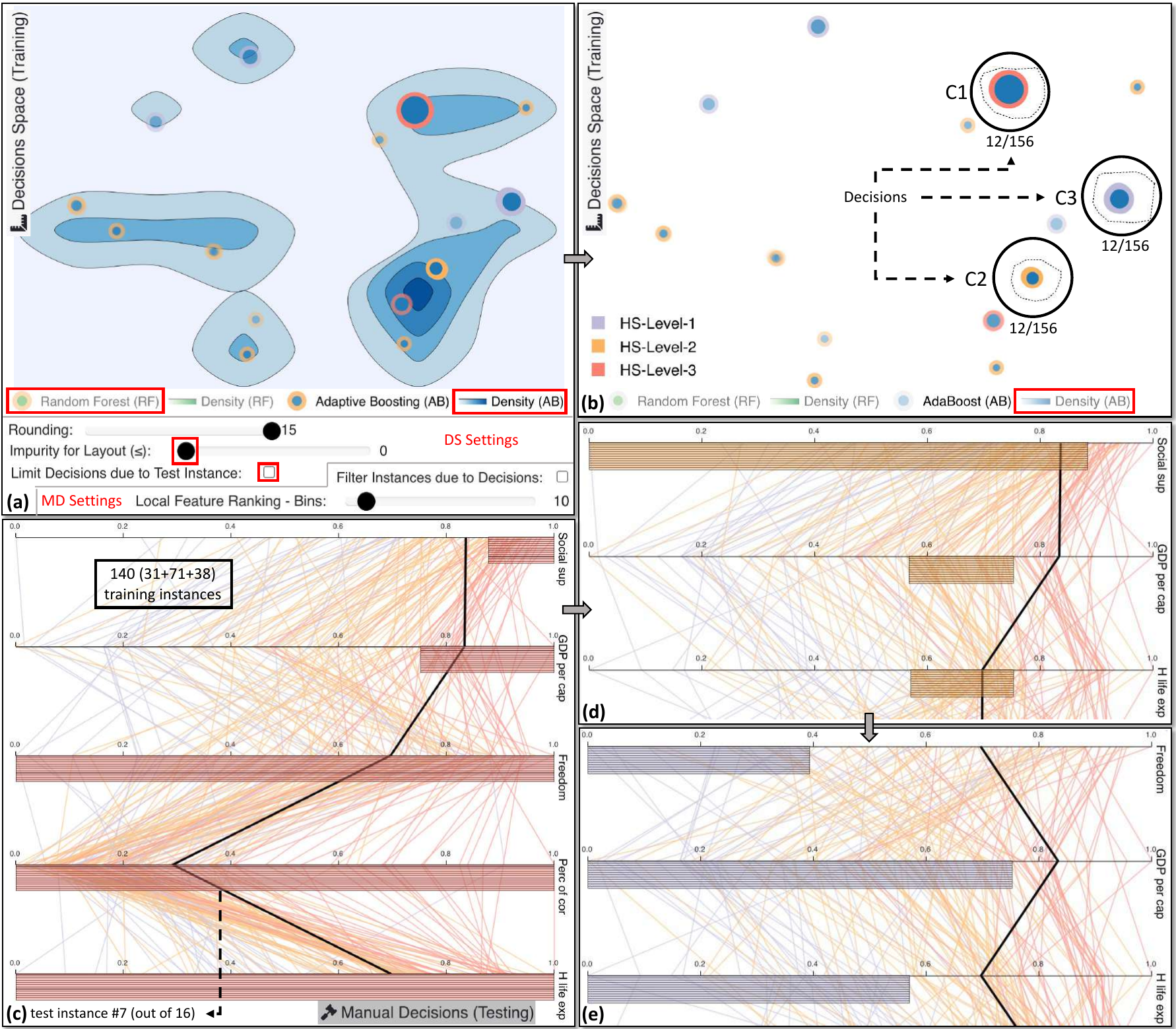}
\caption{Examining several pure global decisions from the active AB model. In (a), we activate the density view in order to distinguish where most decisions are positioned. Note that this screenshot is composed of the \emph{decisions space} (DS) view and the settings for the same view plus the settings for the \emph{manual decisions} (MD) view. In (b), we select step-by-step three clusters of 12 identical decisions each. The decisions for \circled{C1} classify training instances only for HS-Level-3 class (as depicted in (c)). Similarly, \circled{C2} contains decisions for HS-Level-2 (visible in (d)), while \circled{C3} for the remaining class, as shown in (e). The \nth{7} test instance, which is currently under investigation, cannot be classified by those prior decisions. However, it most likely belongs in the medium- or the high-level class.}
\label{fig:use_case1_safe}
\end{figure*}

The vertical PCP-like view in Figure~\ref{fig:teaser}(d) illustrates the range values per feature for each selected decision (comparison mode is active). The vertical polylines represent the training instances and are color-encoded based on the ground truth (GT) class. There are two options: either select to filter instances and show those that belong to the selected rules (see~Figure~\ref{fig:teaser}(d)) or present all training instances at once (see Figure~\ref{fig:use_case1_safe}(c)--(e)). For example, in Figure~\ref{fig:use_case1_safe}(c), we see 12 identical rules that classify the training instances in the HS-Level-3 class (the red horizontal lines). The thick black polyline is the currently explorable test instance; users can compare it to the training instances of the models. All ranges for the features are normalized from 0.0 to 1.0. Scrolling is implemented when many decisions must be shown or the number of features is large. 
The order of the features is initially the global one, as described in Section~\nameref{sec:features}. When a group of points is selected using the lasso tool in the \emph{decisions space} (DS) view, a contrastive analysis~\cite{Zou2013Contrastive} is used to rank the features and highlight unique features that explain a cluster's separation from the rest. The computation works as follows:
(1) break each feature into two disjoint distributions: the values inside the selected group vs. all the rest of the points; 
(2) discretize the two distributions of each feature into bins based on the \emph{Local Feature Ranking - Bins} value set by the user (default is 10); 
(3) compute the cross-entropy~\cite{Mannor2005The} between the two distributions of each feature: higher values of cross-entropy suggest more unique features (i.e. the within-selection distribution is very different than the rest), while lower values suggest more common, shared features; and
(4) rank the features based on step 3, with the more unique features near the top. Either with the overlap or difference setting selected as discussed in Section~\nameref{sec:space}, the decision ranges bounding each feature are visualized in the vertical PCP with a magenta color in this comparison mode.

\subsection{Decisions Evaluation} \label{sec:evaluation}

The panel in Figure~\ref{fig:teaser}(e) contains interactive views that help users find outliers, borderline cases, and misclassified cases in the test set. The first main view supports extracting the \emph{manual decisions} (MD) from the previous phase (see Section~\nameref{sec:manual}). This output is stored in a JSON format where the boundaries of values per-feature are observable for every picked decision rule. This enables domain experts to reuse the hand-picked decision rules for supporting future actions, and also helps in concentrating on cases where the majority of the RF and AB models disagreed when compared to the GT, or for models that did not vote unanimously. It is also possible to go through all test instances one by one. The class agreement between RF and AB models, MD, and the GT is demonstrated via a horizontal stacked bar chart. The colors encode the different classes, and the length of each bar is the number of decisions for (1) MD, (2) RF models, (3) AB models, and (4) the GT (the latter always fills the entire bar). 
The second main view is used to train new models based on the \emph{Ov. Score (\%)} of each previously-trained model. The two separate standard PCPs present the active RF models in green and the active AB models in blue. The brown color is used for the inactive models. 

\subsection{Use Case} \label{sec:application}

In our use case, we observe that models with ID number 8 and above slightly outperform the rest; notably, recall in AB7 is much lower than AB8 and beyond (cf.~Figure~\ref{fig:use_case1_model}(a), line chart). While RF models perform consistently better than AB models, as shown in both the line chart and the confusion plot of~Figure~\ref{fig:use_case1_model}(a), there is an improvement in the score of AB10. Therefore, we decide to keep only this model. Furthermore, since RF8 is more reliable in training instances for the HS-Level-2 class due to false-positives being lower than the equivalent for RF9 and RF10 (Figure~\ref{fig:use_case1_model}(a), confusion plot), we keep this model and RF10, i.e., the top-performing model of the RF algorithm. In consequence, RF8, RF10, and AB10 are active models after selecting the corresponding states. 

At this point, we want to investigate which features of the training set impacted the predictions more (see Figure~\ref{fig:use_case1_model}). Interestingly, \emph{GDP per cap}, \emph{H life exp}, and \emph{Social sup} are the top three features in the general ranking, as in Neto and Paulovich.~\cite{Neto2021Multivariate} A surprising outcome is that, although two of the features mentioned above are still the most important for the selected RF models (all except Social sup), this is not true for the AB model. As seen in Figure~\ref{fig:use_case1_model}(b), \emph{Social sup}, \emph{Perc of cor}, and \emph{GDP per cap} are vital features for the AB algorithm in general. This pattern supports our hypothesis that different algorithms might take into account alternative features and should be combined to provide a holistic view. On the contrary, \emph{Generosity} is unimportant for all models, specifically for the active models, since there is a $-0.12$ decrease in importance. Thus, we choose to remove this feature and retrain without it (cf. Figure~\ref{fig:use_case1_model}(b)). For the RF algorithm (green), we pick the most performant models based on the overall score (Figure~\ref{fig:use_case1_model}(c)), rightmost models). However, AB8 is better overall than the subsequent AB models due to the stable and high recall value (Figure~\ref{fig:use_case1_model}(c), line chart). In a one-to-one comparison between RF9 and AB9 with the bar charts, we recognize that while they have the same \emph{number of estimators} (i.e., 17 trees), the two models produce 555 and 238 decisions, respectively. In this case, bagged decision trees allow a higher maximum depth than the equivalent boosted decision trees. After the selection of the new models, the most important features collectively are \emph{H life exp} with $0.72$ and \emph{GDP per cap} with $0.49$, as illustrated in Figure~\ref{fig:use_case1_model}(d); the opposite was valid in Figure~\ref{fig:use_case1_model}(b). The new AB model considers the same features more important as the RF models. After this phase is over, AB8, RF9, and RF10 are the remaining three active models.

\begin{figure*}[tb]
\centering
\includegraphics[width=\linewidth]{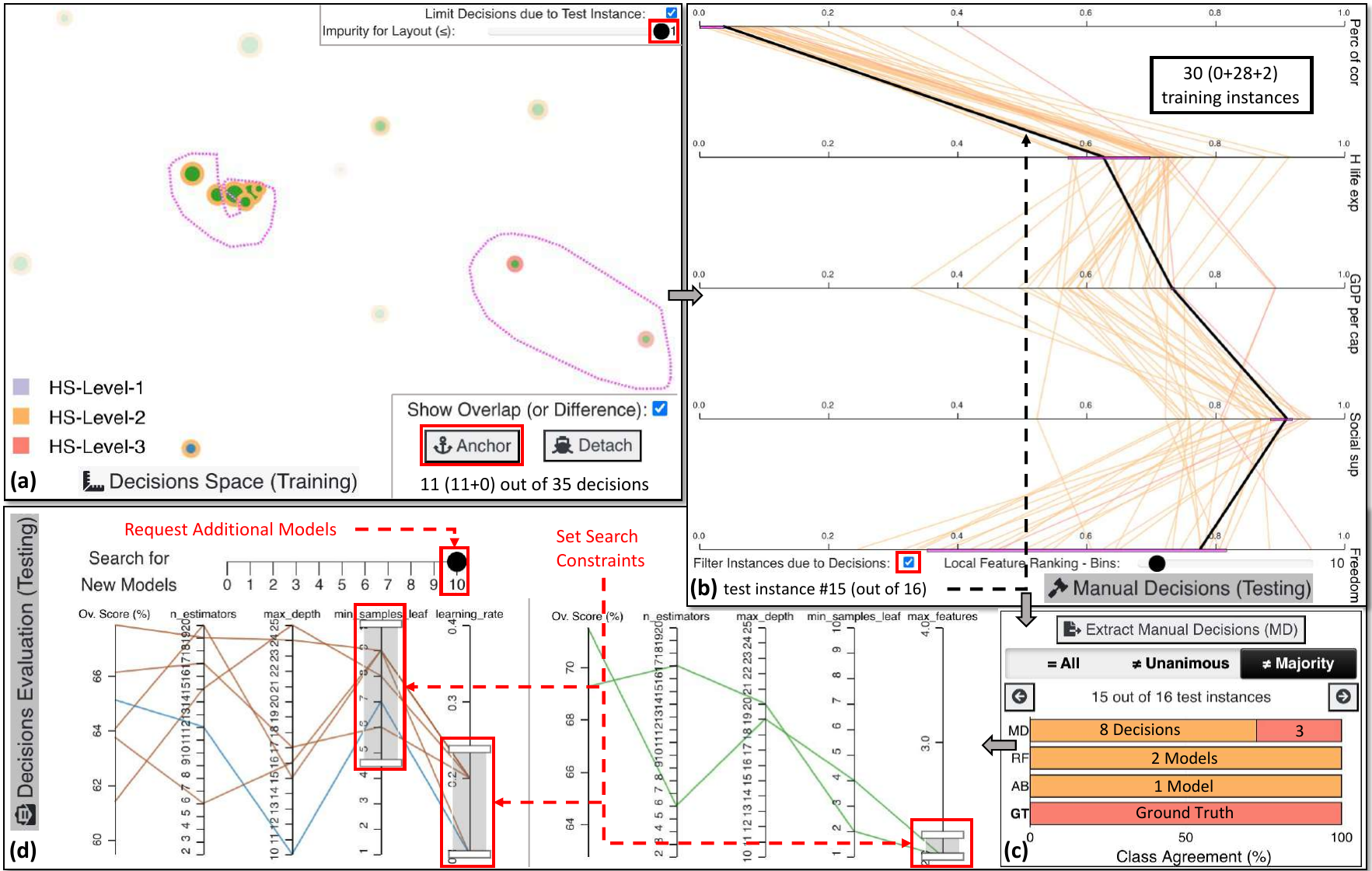}
\caption{An outlier case exploration, the final prediction, and the training of another bunch of RF and AB models. (a) presents the anchoring of a cluster of 8 HS-Level-2 decisions to compare the overlapping rules against 3 HS-Level-3 decisions. In (b), after checking the common regions of agreement for the two clusters, we conclude that \emph{Perc of cor} and \emph{H life exp} are relatively low for the \nth{15} test instance to belong in HS-Level-3 class. However, the other values for the remaining features are arguably rather high. In (c), we observe that all models voted for the average class while only the 3 selected manual decisions are supporting this case to be categorized as HS-Level-3 country. (d) showcases a potential search for new models by setting constraints in the hyperparameters according to the knowledge acquired from the initial training.}
\label{fig:use_case1_outlier}
\end{figure*}

To investigate the global decisions based on the AB8 model we set the impurity to 0, disable limiting decisions based on the current test instance, hide the RF models, and reveal the density view of the active AB model (cf. Figure~\ref{fig:use_case1_safe}(a)). Most decisions are positioned in the right-hand side of the projection. Thus, we continue with the exploration of identical pure decisions from that region. After hiding back the Density (AB) as shown in Figure~\ref{fig:use_case1_safe}(b), we notice from the size of the decisions that if we analyze three core clusters (\circled{C1}--\circled{C3}) we can get a better understanding of global decisions. In Figure~\ref{fig:use_case1_safe}(c), all 140 training instances ($31+71+38$ spread across the classes) are observable together with the \nth{7} test instance, which is currently under investigation because the majority of the RF and AB models disagree with the GT \hl{(not shown due to space limits, but a similar case is visible in Figure~\ref{fig:teaser}(e))}. From Figure~\ref{fig:use_case1_safe}(c), we see that \emph{Social sup} and \emph{GDP per cap} should be very high for test instances to belong to this class. In contrast, for test instances to be in the HS-Level-2 class, they need to have a low-to-average \emph{Social sup}, and average \emph{GDP per cap} and \emph{H life exp} (Figure~\ref{fig:use_case1_safe}(d)). Low values in the features (1) \emph{Freedom}, (2) \emph{GDP per cap}, and (3) \emph{H life exp} are common for the low score in happiness countries (see Figure~\ref{fig:use_case1_safe}(e)), as also identified by Neto and Paulovich.~\cite{Neto2021Multivariate} Regarding Saudi Arabia (the \nth{7} test instance), it does not appear to belong to any of those decisions, but it is far away from the values reported for the HS-Level-1 class. It has a very high \emph{GDP per cap} to belong in the average class, but the \emph{Social sup} is on the lower side. Despite that, \emph{GDP per cap} is 1 out of the 2 most important features according to the analysis in Section~\nameref{sec:features}. Our conclusion matches the fact that it was ranked in \nth{28} place out of the 156 countries, thus, belonging to the list of 42 countries classified as HS-Level-3.

Using the tool's mechanism to detect problematic test instances, a borderline case that stands out is the \nth{6} test instance. in Figure~\ref{fig:use_case1_broderline}(a), we first lower the impurity threshold to focus only on completely pure decisions. This example is for a specific case, thus, \emph{Limit Decisions due to Test Instance} is checked. \circled{C1} seems to have a zero impurity, but when zooming in, we recognize that it contains several decisions with very high impurity values. Hence, we ignore this cluster, and we move to \circled{C2} with a decision influencing only 2 training instances. Since this number is undoubtedly low and could overfit these 2 instances, we increase the lower limit of visible decisions through the bar chart at the bottom. We exempt 4 decisions with 2 or 3 influenced training instances spread throughout the projection with this action. An interesting insight is observable in \circled{C3}: 8 decisions produced by the RF models are divided, with decisions suggesting that the test instance should be classified as either HS-Level-1 or HS-Level-2. We anchor one of the subclusters and select the other for comparison by viewing the difference in the value ranges of the five features (cf. Figure~\ref{fig:use_case1_broderline}(b)). 12 out of the 18 training instances suggest that Guinea (i.e., the \nth{6} instance) is similar to low-happiness countries. If \emph{Perc of cor}, \emph{H life exp}, and/or \emph{GDP per cap} were slightly higher, then the outcome would have been entirely different. According to the GT ranking Guinea is in \nth{118} place, remarkably close to the \nth{122} test instance which is the first country classified as low happiness.

\begin{figure*}[tb]
  \centering
  \includegraphics[width=\linewidth]{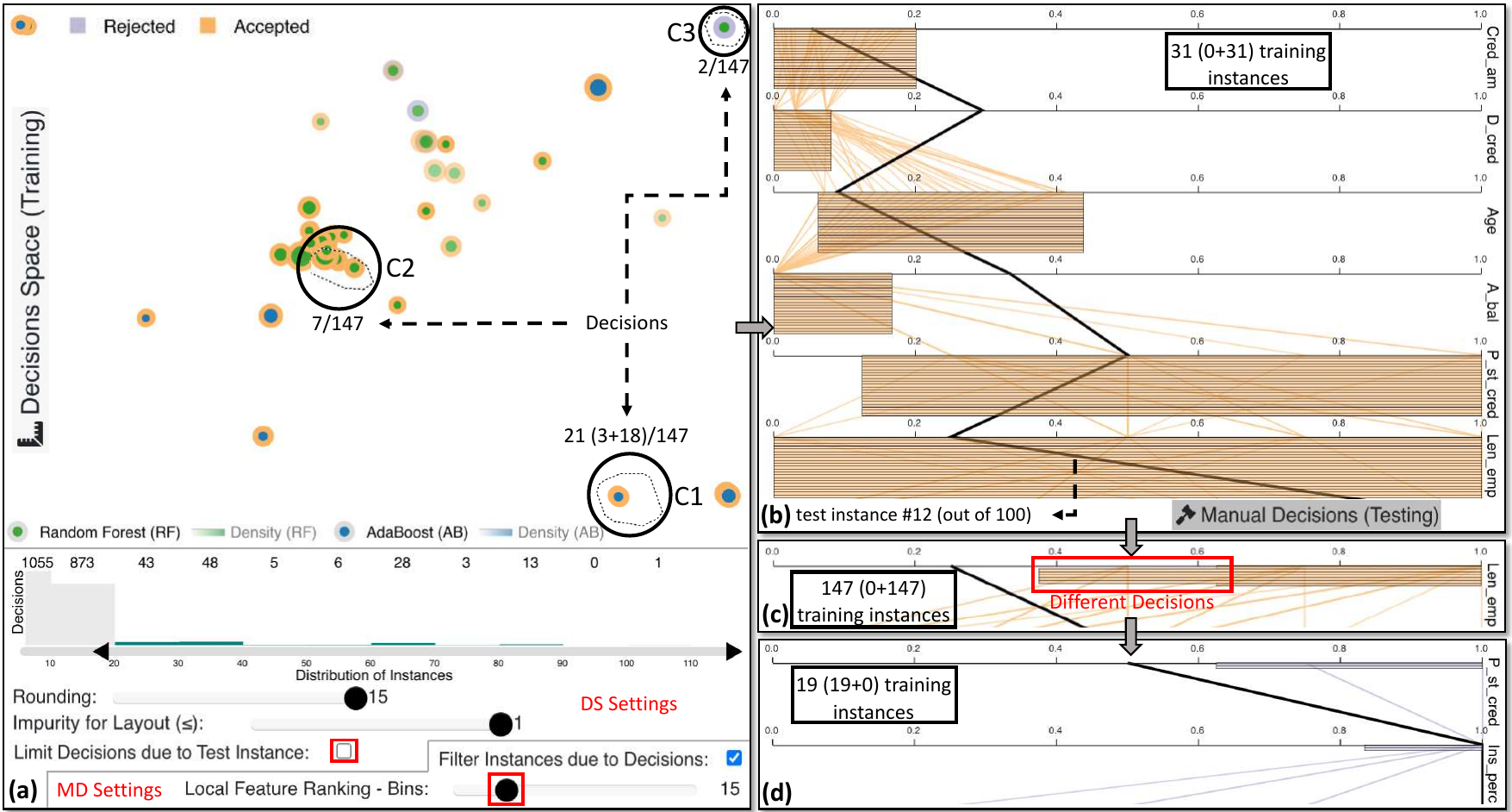}
  \caption{The exploration of clusters of decision paths from both ML algorithms. View (a) presents the selection of three clusters of global decisions that classify multiple training instances, thus, avoiding unimportant paths that might overfit. (b) provides an in-depth analysis of the decisions rules affected by \circled{C1}. In (c), \emph{Len\_emp} emerges as a unique feature that characterizes \circled{C2} with values from approximately 0.4 to 1.0. Finally in (d), high values in \emph{P\_st\_cred} and \emph{Ins\_perc} turn over the prediction of the applicant to reject, visible via the exploration of \circled{C3}.}
  \label{fig:use_case2_safe_global}
\end{figure*}

Checking the cases where the majority of the models disagree with the GT, we stop in the \nth{15} test instance. Figure~\ref{fig:use_case1_outlier}(a) shows the decisions applicable for this unusual case. We use the comparison mode to select a pure cluster on the left to juxtapose it with decisions classifying countries as HS-Level-3 on the right. Anchoring these clusters of points shows us the overlap of value ranges for the different features, as depicted in Figure~\ref{fig:use_case1_outlier}(b). 28 out of the 30 training instances are similar to this test instance and belong to the HS-Level-2 class. The ranking of the features indicates that \emph{Perc of cor} and \emph{H life exp} are two unique features for the selected points, with low values for the former and average values for the latter, as in Neto and Paulovich.~\cite{Neto2021Multivariate} Furthermore, for the first four features, the overlap is narrow between the two selected clusters, indicating that this instance could be considered an outlier. Indeed, Figure~\ref{fig:use_case1_outlier}(c) presents that 8 out of the 11 decisions consider this instance as HS-Level-2. All active models are wrongly predicting Trinidad and Tobago (i.e., the \nth{15} test instance) as an average HS country. Interestingly, the 3 MD of the RF models classified this country as HS-Level-3.

From the analyses and the overall score of the RF and AB models, we observe that the most performant models for RF consider only 2 features when splitting the nodes (i.e., \emph{max\_features} hyperparameter). The PCPs in Figure~\ref{fig:use_case1_outlier}(d) enable us to scan the internal regions of the hyperparameters' solution space for RF. As for AB, the \emph{learning\_rate} should be as low as possible for this specific data set, as seen in Figure~\ref{fig:use_case1_outlier}(d). Also, by searching for models with high values for \emph{min\_samples\_leaf}, AB models are created with complex decision trees compared to simple decision stumps, which seems to be an appropriate limitation of the hyperparameter space that could lead to better models. After all these constraints, we move the \emph{Search for New Models} slider from 0 to 10 in Figure~\ref{fig:use_case1_outlier}(d) to request 10 additional models for each algorithm with the hope of discovering more powerful ones. \hl{In summary, \textsc{VisRuler} supported the exploration of diverse decision rules extracted from two different ML algorithms and boosted the trustworthiness of the decision making process (\textbf{RQ1}).}

\section{Usage Scenario} \label{sec:case}
  \begin{table*}[htb]\centering
\captionsetup{justification=justified}
\caption{User study: Shortened questions (Qs), goals (Gs), the ground truth, and results. There are five multiple choice questions, each with four possible answers. The
goals and the ground truth (GT) can be found in Section~\nameref{sec:goals} and Section~\nameref{sec:overview}, respectively. The results are computed as: number of correct answers / total number of participants.}
\label{results}
\setlength\tabcolsep{0pt} 
\begin{tabular*}{\linewidth}{@{\extracolsep{\fill}} l ccc}
\toprule
\multirow{1}{*}{\textbf{Question (1, 2, 3, 4, \& 5)}} & \multicolumn{1}{c}{\textbf{Goal}} & \multicolumn{1}{c}{\textbf{GT}} & \multicolumn{1}{c}{\textbf{Result}} \\
\midrule
     If you \textbf{must remove a feature}, which one will it be? & \textbf{G2} & \ref{fig:use_case1_model}(b) & 12/12 \\
     Which models present \textbf{stable and high performance} based on \textbf{all} validation metrics? & \textbf{G1} & \ref{fig:use_case1_model}(c) & 9/12 \\
     \textbf{How many test instances are in conflict} and require human intervention (that you can identify)? & \textbf{G5} & \ref{fig:use_case1_outlier}(c) & 9/12 \\
     Which \textbf{features play a vital role} for the classification of instances in the \textbf{HS-Level-3 class}? & \textbf{G3\&G4} & \ref{fig:use_case1_safe}(b) & 12/12 \\
     Which \textbf{feature's value is low} and \textbf{not contributing} to the \textbf{\nth{15}} instance's classification as HS-Level-3? & \textbf{G3\&G4} & \ref{fig:use_case1_outlier}(b) & 11/12 \\
\bottomrule
\end{tabular*}
\label{fig:questions}
\end{table*} 

In this section, we describe a hypothetical usage scenario with a collaboration of a model developer (Amy, the ML expert) and a bank manager (Joe, the domain expert) who handles granting loans to customers. Joe wants to use \textsc{VisRuler} to improve the evaluation process of loan requests, so he asks Amy to use \textsc{VisRuler} to train ML models based on a data set collected over years of accepting or rejecting loans in the bank. The data set includes 1,000 instances/customers and 9 features/customer information, with 300 rejected (purple) and 700 accepted (orange) applications. This data set is, in reality, a pre-processed version~\cite{Zhao2019iForest,Neto2021Explainable} of German Credit Data from the UCI ML repository.~\cite{Dua2017}

\textbf{Exploration and Selection of Algorithms and Models.} Following the workflow in Section~\nameref{sec:overview}, Amy loads the data set and checks the score of each model based on the three validation metrics (Figure~\ref{fig:teaser}(a)). For the AB algorithm, in blue, all models have a relatively low value for the recall metric, except for AB8. Also, AB7 performs very well for the Accepted class (orange), since the false-negative (FN) line reduces in height compared to all other models. Therefore, she decides to keep only AB7 and AB8. By looking at the confusion plot in Figure~\ref{fig:teaser}(a), Amy infers that RF5 is the model with low confusion regarding the Rejected class (purple). She is determined to use RF5 because it carries over only 104 different false-positive (FP) instances compared to RF4 with 114. The top RF models on the right-hand side also caught her attention, with RF9 and RF10 being the best options. She thinks that either of them could do the job, as they appear redundant due to similar confusion and values in both the confusion plot and the line chart (cf. Figure~\ref{fig:teaser}(a)). The bar charts below---which highlight the difference in the architectures of these RF models---help her to choose: with only 7 decision trees and 589 decision paths (compared to 18 and 1,483), RF9 is simpler. She concludes that RF9's simplicity will make Joe's exploration of decisions more manageable later. Consequently, she deactivates RF10 and continues the feature contribution analysis with RF5, RF9, AB7, and AB8 models.

\textbf{Examining the Global Contribution of Features.} After this new selection of models, Amy observes in Figure~\ref{fig:teaser}(b) that most features (except for the last two) are more important now than in the initial state. \emph{Ins\_perc} and \emph{Val\_sa\_st} importances drop only by 0.01, implying these features are stable. She suggests Joe to keep all features for now and explore the differences through the decision rules later on. Another interesting insight is that \emph{A\_bal} is the most important feature for the RF models, while the AB models prefer \emph{D\_cred} (see Figure~\ref{fig:teaser}(b)). This could indicate that mixing models' decisions from different algorithms is beneficial. 

\textbf{Explanations through Global Decision Rules.} Joe starts his exploration by examining the global decision rules that can help him make accurate decisions for specific cases in the future. 
He focuses on the \nth{12} test instance, which is a customer application reviewed by a colleague, Silvia (cf. usage scenario by Neto and Paulovich~\cite{Neto2021Explainable}). First, he unchecks \emph{limiting the decisions due to the test instance}, as illustrated in Figure~\ref{fig:use_case2_safe_global}(a). At this point, Amy identifies several decisions that classify only fewer than 20 customers; she thinks: ``these are not so generic after all''. Indeed, the larger the number of instances classified by one rule, the more generic and important it is (if the impurity is low). Consequently, they decide to increase the lower boundary of decisions, filtering out 1,928 decisions (see Figure~\ref{fig:use_case2_safe_global}(a), bar chart). After the update, Joe focuses on the UMAP~\cite{McInnes2018UMAP} projection. He observes multiple groups of points that could be worthy of further investigation. He selects a couple of samples from different areas, e.g., \circled{C1} with 3 RF and 18 AB decisions. Another cluster with 7 decisions is \circled{C2} that solely predicts accepted loan applications. On the contrary, \circled{C3} contains 2 pure decisions (due to high opacity) that produce rules which reject loans. Joe increases the \emph{discretization of local feature ranking from 10 to 15 bins} to raise the sensitivity of difference between decision rule ranges, and he \emph{filters the instances due to the decisions} to observe clearer trends. From Figure~\ref{fig:use_case2_safe_global}(b), Joe recognizes that \circled{C1} decisions are all identical, having the same ranges for every feature. Also, he understands that low \emph{credited amount} (\emph{Cred\_am}) and short \emph{duration of credit} (\emph{D\_cred}) are essential factors for accepting a loan application. \emph{Account balance} is also vital because all loans are accepted when there is no account (\emph{A\_bal} being 0). Figure~\ref{fig:use_case2_safe_global}(c) reveals another intriguing pattern, that is, \emph{the length of current employment} should be average to extremely high (from approximately 0.4 or 0.6 and above\hl{, shown in the red box}) for applications to get accepted. In contrast, Figure~\ref{fig:use_case2_safe_global}(d) presents that if \emph{payment status of previous credit} (\emph{P\_st\_cred}) and \emph{instalment per cent} (\emph{Ins\_perc}) are relatively high, the applications were rejected. The \nth{12} customer has an account without any balance, and the \emph{D\_cred} is relatively high, which flips the prediction toward rejection. Luckily, Silvia also provided an adequate justification to the customer.~\cite{Neto2021Explainable}

\textbf{Extracting Manual Decisions through Local Investigations.} At this point Joe knows and understands the main decision rules, but a new customer arrives. 
Focusing on the decisions for this case (i.e., \nth{90} test instance), he sets impurity to less than 0.3 (cf. Figure~\ref{fig:teaser}(c), slider) to make impure decisions more transparent. Two fairly pure decisions from RF5 (visible due to hovering) and RF9 contradict each other. Joe uses the comparison mode, anchors 1 out of the 2 decisions, and selects the other with the lasso tool. The comparison in Figure~\ref{fig:teaser}(d) designates that 8 similar customers' applications were rejected while 12 were accepted. The small overlap in \emph{Cred\_am}, \emph{D\_cred}, and \emph{Age} suggest that this is a borderline case. \emph{Cred\_am} seems a bit arbitrary for the training data since only a small amount of applications in-between accepted applications were rejected, see Figure~\ref{fig:teaser}(d), feature on top. However, a clear insight is that if \emph{D\_cred} was lower, the application should have been accepted, while the opposite effect is true if the \emph{duration of credit} increases. Unexpectedly, RF models vote for accepting this loan application while AB models reject (cf. Figure~\ref{fig:teaser}(e), top view). Besides that, the manual decisions are also in-between the two classes, which further enhances Joe's assumption that this is a borderline case. As AB models propose rejection and RF9 produces a decision for rejecting this application, he follows these recommendations. Nonetheless, Joe asks Amy to search and train new performant ML models (see next paragraph).

\textbf{Tuning the Search for Bagged and Boosted Decision Trees.} Amy sees two possibilities of improvement for the RF in Figure~\ref{fig:teaser}(e), bottom view. One is to limit the \emph{max\_features} to 7 because it produces the two best models so far (visible by following the lines at the very top in \emph{Ov. Score (\%))}. The second strategy is to pick 3 and 4 for the same hyperparameter to explore an entirely new space of currently unexplored models since there is no existing line. Basically, she believes it is better to try both strategies in two separate runs. As for the AB, she reasons that selecting 0.1 and 0.2 for the \emph{learning\_rate} is a wise choice. Although it may take more time to retrain the AB models, they probably will be more powerful than with the other setting due to historical data. She performs the above actions, and finally, another cycle of exploration is unfolded for both experts. \hl{To summarize, our VA system not only helps users to reason about concrete cases, but also is capable of assisting ML experts and domain experts in enhancing their overall understanding due to their collaboration throughout the entire process (\textbf{RQ2}).}

\section{User Study} \label{sec:eval}
  We conducted a user study to evaluate our tool's effectiveness in supporting decision-making. 
As in prior works,~\cite{Ming2019RuleMatrix,Neto2021Explainable} we created five questions (Qs) 
that cover \textsc{VisRuler}'s different views,
focusing on appraising the goals described in Section~\nameref{sec:goals} with the use case outlined in Section~\nameref{sec:overview} as the GT (see Table~\ref{fig:questions}). Note that this user study was based on a slightly different arrangement of visualizations for the \emph{models overview} panel, but all the rest of the visual representations and in-depth functionalities remained the same. In particular during the study, the line chart and the confusion plot were not aligned with the bar charts below, and the legends of this panel were positioned at the very top instead of beside the visualizations (as in Figure~\ref{fig:teaser}(a)).

\begin{figure}[h]
  \centering
  \includegraphics[width=\linewidth]{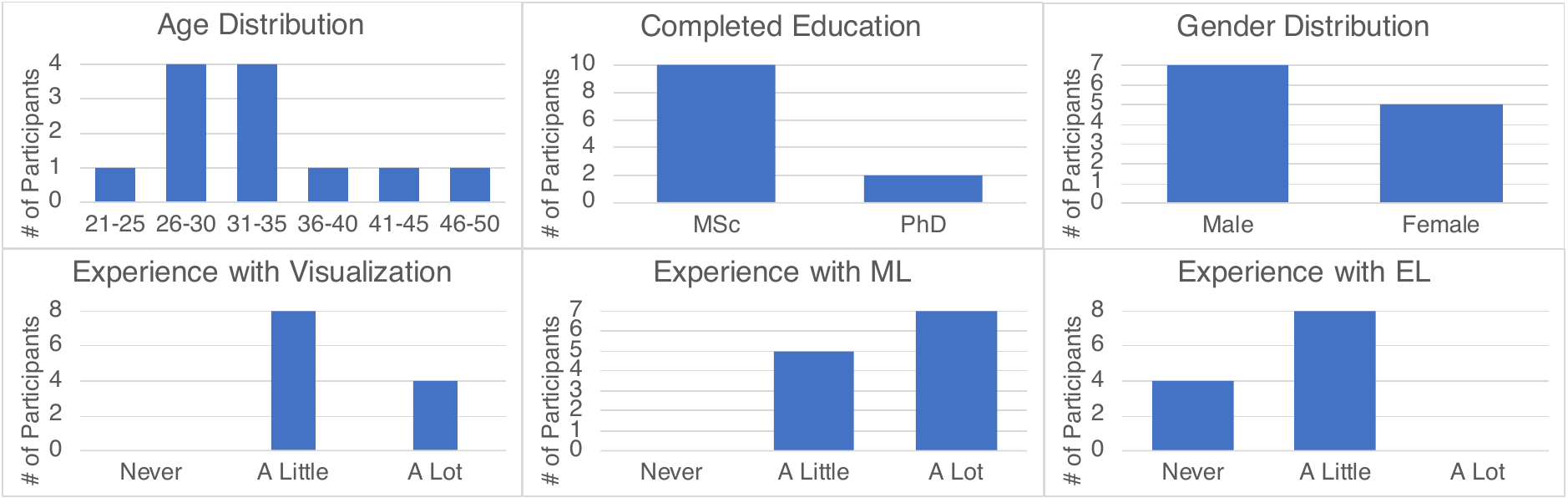}
  \caption{General information on the participants of our user study.}
  \label{fig:demographics}
\end{figure}

\textbf{Demographics.} Figure~\ref{fig:demographics} contains general information about the attendees of the user study. Seven male and five female volunteers aged 23 to 49 (mean: $\approx$33) participated in our study, all with at least an MSc degree (and two PhD's). None of them knew the data set used, and no colorblindness issues were reported. Four of the participants were highly knowledgeable in visualization and seven in ML, while the rest had limited knowledge regarding all aspects. Additionally, four of them had never worked with any EL method. All participants were researchers that have to frequently work with ML and tune ML models for various data tasks, such as image classification and natural language processing.

\textbf{Methodology and Instructions.} Initially, participants watched an $\approx$18-minute video tutorial about bagging and boosting concepts, \textsc{VisRuler}'s goals, and how to work with our tool to analyze decision paths, using the Iris data set.~\cite{Fisher1936The} The participants experimented for five minutes with Iris, which concludes the required training for utilizing the capabilities of our tool. Then, they proceeded to use the data set described in Section~\nameref{sec:overview}. They were asked to answer five questions (cf. Table~\ref{fig:questions} for a summary). The original document containing the questions and instructions as shown by the attendees is available in the supplemental material accompanying this paper. Finally, the participants were requested to provide qualitative feedback via the ICE-T questionnaire.~\cite{wall2019aheuristic}
%

\begin{figure}[h]
  \includegraphics[width=\linewidth]{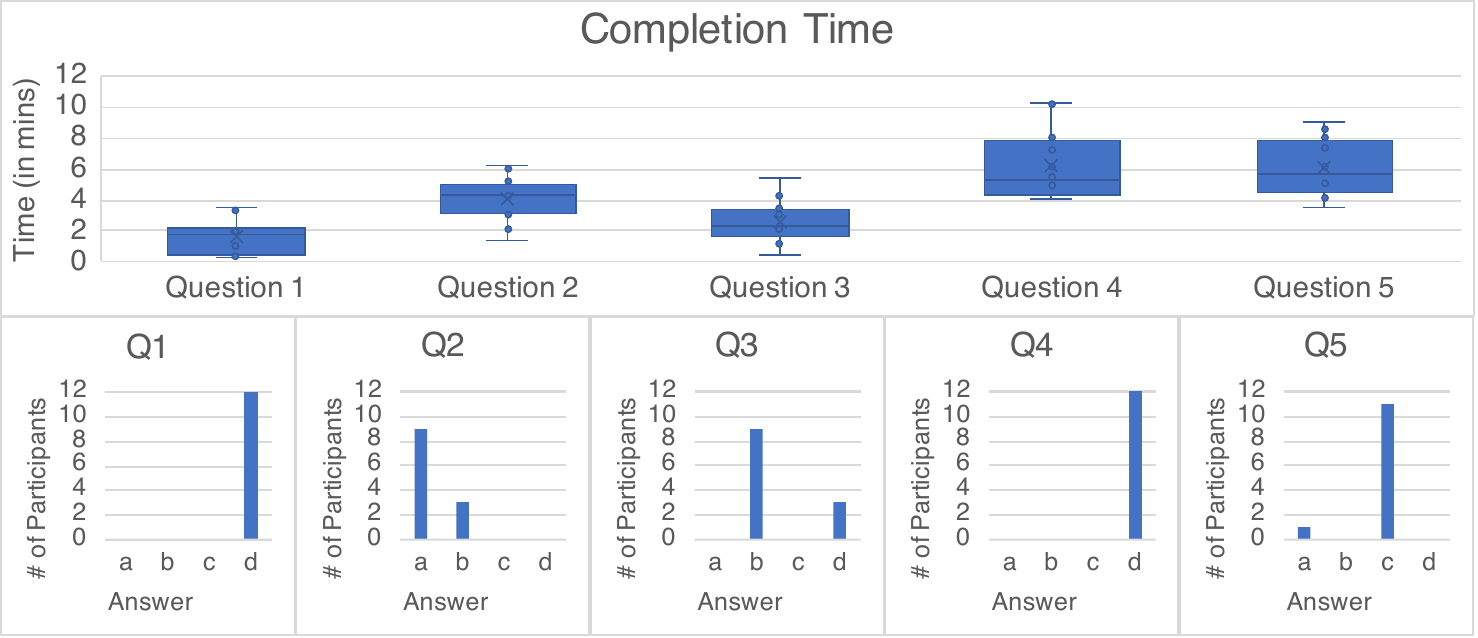} 
  \caption{The question-related results of the user experiment. The top row presents the \emph{completion time} for every question of the study separately, and the bottom row comprises the histograms of the participants' answers in all questions.}
  \label{fig:results}
\end{figure}

\textbf{Question-related Results.} The completion time it took the users to respond to each question and their answers to the multiple choice questions are shown in Figure~\ref{fig:results}. After the initial setting shown in Figure~\ref{fig:use_case1_model}(a), all participants decided to exclude \emph{Generosity} in Q1 (Answer: d, Q1), which happened in 2.03 minutes on average. For Q2, 9 participants followed our GT (Answer: a, Q2), as described in Figure~\ref{fig:use_case1_model}(c). The remaining attendees selected AB10 instead of AB8 (Answer: b, Q2). This action led to 5 test instances in conflict (Answer: b, Q3) compared to 3 (Answer: d, Q3) in our analysis (Figure~\ref{fig:use_case1_outlier}(c) presents a single case). This result could be a strong indication that our approach is essential for making such decisions. To respond in Q2 and Q3, participants took 4.04 and 2.58 minutes on average, respectively. The most time-consuming question was Q4 with an average response time of 6.15 minutes (but with very accurate results (Answer: d, Q4), see  Figure~\ref{fig:use_case1_safe}(b)).
%
The average time taken for Q5 was 6.07 minutes, with all correct answers (Answer: c, Q5) except for one (Answer: a, Q5). The participant that responded incorrectly chose \emph{Freedom} because it was at the bottom of the vertical PCP (cf. Figure~\ref{fig:use_case1_outlier}(b)).

\textbf{Qualitative Results.} 
In Table~\ref{fig:ICET}, the mean scores of the ICE-T components~\cite{wall2019aheuristic} for each participant are displayed along with the two-tailed 95\% confidence intervals (CIs) per component ($t^* = 2.201$, $N=12$). Higher values in green indicate good results, as opposed to red. \textsc{VisRuler} has received a few 7.0 scores, and most are at least 6.0 and above (the lowest score is 4.67). Essence, Insight, and Time received large scores which means users found our tool competent in portraying decisions, guiding users to come up with fundamental questions, and performing these discoveries quickly. Confidence was lower, with a mean value of 5.87. However, this value still makes \textsc{VisRuler} a reliable and trustworthy VA tool according to Wall et al.~\cite{wall2019aheuristic}

\begin{table}[h]
\captionsetup{justification=justified}
\centering
\caption{Analyzed results from the ICE-T feedback.~\cite{wall2019aheuristic}}
\includegraphics[width=\columnwidth]{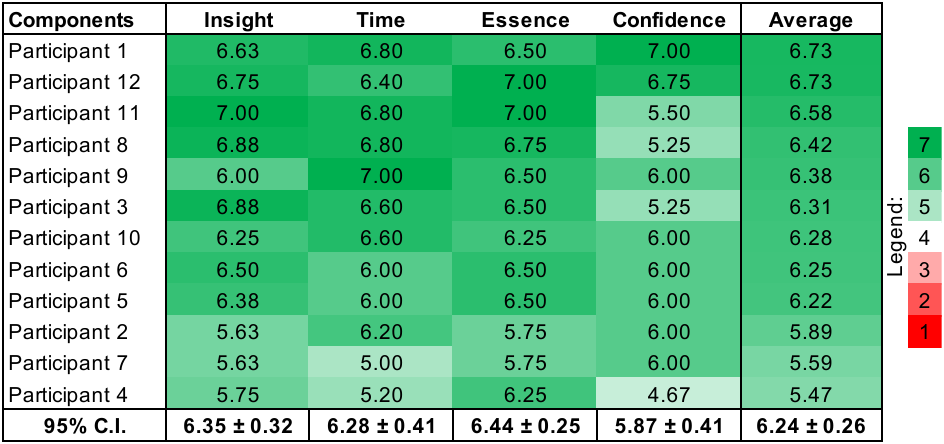}
\label{fig:ICET}
\end{table}


\section{Limitations and Future Work} \label{sec:lim}
  In this section, we discuss several limitations of our VA system that could be regarded as improvement ideas for the future.

\textbf{Generalization.} Since \textsc{VisRuler} mainly focuses on the exploration of decision paths, our approach is applicable to any tree-based algorithm. If we follow the same methodology, RF could be changed to extremely randomized trees~\cite{Geurts2006Extremely} (known as extra trees) or other bagging algorithms; while instead of AB, gradient boosting~\cite{Ke2017LightGBM,Chen2016XGBoost} or any boosting algorithm can be employed. An extension of our approach could be to include new supervised ML methods such as rule-based algorithms, or even association rule learning.~\cite{Song2016Research} However, such algorithms consist of two parts: the antecedent (IF) and the consequent (THEN), which are complicated because there could be multiple if statements that bind many times the values using the same features. Therefore, a single decision path is non-trivial to be extracted, as in our case. One potential enhancement would be to support rule-based algorithms to encode these reoccuring information, which implies an order of consecutive events. 

As shown in the paper, \textsc{VisRuler} works with both binary and multi-class classification problems. Since humans may have problems in perceiving more than ten categorical colors~\cite{Ware2019Information} at the same time, \textsc{VisRuler} utilizes all of them as well as possible. Nonetheless, a limitation is the extensive (but unavoidable) use of color that might hinder our tool from operating with more than a few classes. Therefore, applying the one-vs-rest strategy is one possible idea to solve multi-class classification problems with several classes. This strategy translates to either designating one class as the positive class and all others as the negative class or choosing classes and gradually examining them.

At last, although \textsc{VisRuler} concentrates on the interpretation of rules and extraction of manual decisions driven by the experienced domain experts, it can also be used by ML experts to tune the hyperparameters of models and eventually debug RF and AB models. We acknowledge that \textsc{VisRuler} takes premature steps toward this direction, but this concept appears an exciting research opportunity for the future.

\textbf{Scalability.} Similar to many VA tools/systems,~\cite{Robertson2009Scale} a major challenge we considered when designing \textsc{VisRuler} is scalability.
In the \emph{models overview} panel, the number of trained models is limited to 20 RF and AB models in total (or 10 for each algorithm). However, in the backend users can set different hyperparameters and perform hyperparameter search with various automatic hyperparameter tuning approaches~\cite{Claesen2015Hyperparameter,Claesen2014Easy} or VA tools for this same goal.~\cite{Li2018HyperTuner,Chatzimparmpas2021VisEvol} \textsc{VisRuler} utilizes Random search for this purpose due to several benefits identified by Bergstra and Bengio.~\cite{Bergstra2012Random} The end result is to visualize several robust and diverse models in our tool, which can be deemed an adequate number of models. Also, the two PCPs for training new RF and AB models allow users to explore more models progressively, especially since the provided hyperparameters' ranges are also easily modifiable through the code.

In the \emph{decisions space} view, circles may overlap when the number is large, as with any dimensionality reduction technique presented in a scatter plot. Although \textsc{VisRuler} can visualize thousands of decision paths (illustrated by the usage scenario in Section~\nameref{sec:case}), the cluttering of the low-dimensional embedding could be considered as an intrinsic difficulty. To address this issue, we first adopt a filtering approach to remove irrelevant or even overfitting decisions (e.g., see Figure~\ref{fig:use_case1_broderline}(a)), and second we enable users to partially hide out impure decisions (cf. Figure~\ref{fig:use_case1_safe}(a)). Users may also utilize interactions like panning and zooming to focus on certain regions of circles. A potential future idea is to enhance the current scatter plot with approaches designed to reduce overlapping.~\cite{Mayorga2013Splatterplots,Hilasaca2019Overlap}

In the \emph{manual decisions} view while trying to get an overview, the vertical PCP may be challenging to interpret because it requires users to scroll through a list of decisions that expands by the number of features. Despite that, we have implemented multiple layout treatments (e.g., filtering out decisions visible in Figure~\ref{fig:use_case1_broderline}(b)) and interaction possibilities (e.g., the comparison mode for juxtaposing groups of decisions shown in Figure~\ref{fig:use_case1_outlier}(b)) for users to partially overcome these challenges. Furthermore, the most common scenario is to explore regions of decisions paths and focus on specific test instances which by default drastically limits the number of decision rules. In summary, the benefits of this tweaked visualization are many, since users can directly compare a test case with training instances for the various rules applicable and all features of a data set. The vertical PCP can help domain experts to externalize their domain knowledge because it serves as a root for a discussion between experts and the general public. One potential update is to try out alternative PCP designs that could boost the scalability of this view, such as the proposal from Wu et al.~\cite{Wu2017Making}

\textbf{Efficiency.} The performance of \textsc{VisRuler} could pose problems if numerous models are simultaneously active and produce too many decisions. Indeed, the excessive computational time required for exploring thousands of decisions paths along with the initial training of those ML algorithms can be a root cause for further troubles. Using distributed computation processes on performant cloud servers can be one solution for scaling \textsc{VisRuler} to enormous data sets. In the future, we believe that the improvement in high-performance hardware as well as progressive VA approaches~\cite{Stolper2014Progressive,Turkay2018Progressive} will also benefit \textsc{VisRuler}.

The use case, usage scenario, and user study were performed on a MacBook Pro 2019 with a 2.6 GHz (6-Core) Intel Core i7 CPU, an AMD Radeon Pro 5300M 4 GB GPU, 16 GB of DDR4 RAM at 2667 Mhz, running macOS Monterey, and with Chrome (version 99) as the browser. The system can perform interactively after the model training stage is over, which may take a few minutes for the data sets used in the use case of Section~\nameref{sec:overview} and the usage scenario of Section~\nameref{sec:case}. However, we cache the results to speed-up drastically the future executions for the same data sets.

\textbf{Complexity.} Compared to several VA tools/systems,~\cite{Chatzimparmpas2020The} \textsc{VisRuler} is no exception in terms of the high cognitive load that could overwhelm users. Despite that, the proposed workflow of \textsc{VisRuler} (see Figure~\ref{fig:workflow-diagram} and read Section~\nameref{sec:overview}) is mainly linear. Furthermore, the participants of our user study (cf. Section~\nameref{sec:eval}) correctly performed most of the provided tasks after a specific training period, which is indicative of the gradual learning curve of our tool. However, in a future iteration of the tool, we plan to implement a hiding functionality for the \emph{models overview} panel after the initial phase, involving mainly an ML expert selecting powerful and diverse models, is over. Another incremental improvement could be to increase the size of the symbols for the three validation metrics present in the line chart of Figure~\ref{fig:teaser}.

\textbf{Evaluation.} While we already conducted a task-based user study with 12 participants that tested the applicability and effectiveness of \textsc{VisRuler}, additional review sessions with experts could help us to validate our tool further. However, as illustrated in Figure~\ref{fig:collaboration-diagram}, our VA system is designed to be operated with a single workflow for two experts that most of the time are set apart and work independently. The prior knowledge and expertise of each group of experts is useful in specific steps of the collaboration schema, especially since they meet only in step 4, related to the \emph{decisions space} exploration. A threat in this case is the overconfidence effect and overinterpretation of the models' capabilities by both domain-specific and ML experts, especially in noisy data scenarios. Despite that, we believe our first user study was an appropriate choice of method to understand preliminarily if \textsc{VisRuler} is usable and effective. In the future, we could further evaluate the particular designs of this multi-component system with both ML and domain experts.

\section{Conclusions} \label{sec:con}%
  We presented \textsc{VisRuler}, a VA tool that allows users to explore diverse rules extracted from bagged and boosted decision trees to reach a consensus about a final decision for each individual case. The multiple coordinated views facilitate the selection of diverse and performant models, the characterization of per-feature contribution, the management of multiple decisions, the analysis of global decisions, and support case-based reasoning. A use case and a usage scenario with real-world data sets emphasize the necessity for combining similar, but yet, different algorithms and the importance of transparency in critical domains. We also validate the usability and efficacy of \textsc{VisRuler} via a user study. Finally, we describe a series of limitations of our VA system with an ultimate goal to extract future directions for our work.

\begin{acks}
This work was partially supported through the ELLIIT environment for strategic research in Sweden.
\end{acks}

\bibliographystyle{SageV}
\bibliography{references.bib}

\end{document}